# Attributed Graph Clustering with Multi-Scale Weight-Based Pairwise Coarsening and Contrastive Learning

***Binxiong Li[a], Yuefei Wang[b,*], Binyu Zhao[a], Heyang Gao[a], Benhan Yang[a], Quanzhou Luo[a], Xue Li[b], Xu Xiang[b], Yujie Liu[a], Huijie Tang[a]***

[a]*Stirling College, Chengdu University, 2025 Chengdu Rd., Chengdu, Sichuan, China, 610106*

[b]*College of Computer Science, Chengdu University, 2025 Chengdu Rd., Chengdu, Sichuan, China, 610106*

E-mail Address:

libinxiong@stu.cdu.edu.cn (Binxiong Li); wangyuefei@cdu.edu.cn (Yuefei Wang); gaoheyang@stu.cdu.edu.cn (Heyang Gao); zhaobinyu@stu.cdu.edu.cn (Binyu Zhao); yangbenhan@stu.cdu.edu.cn (Benhan Yang); luoquanzhou@stu.cdu.edu.cn (Quanzhou Luo); lixuexue@stu.cdu.edu.cn (Xue Li); xiangxu@stu.cdu.edu.cn (Xu Xiang); liuyujie260@stu.cdu.edu.cn (Yujie Liu); tanghuijie@stu.cdu.edu.cn (Huijie Tang)

**ABSTRACT**

This study introduces the Multi-Scale Weight-Based Pairwise Coarsening and Contrastive Learning (MPCCL) model, a novel approach for attributed graph clustering that effectively bridges critical gaps in existing methods, including long-range dependency, feature collapse, and information loss. Traditional methods often struggle to capture high-order graph features due to their reliance on low-order attribute information, while contrastive learning techniques face limitations in feature diversity by overemphasizing local neighborhood structures. Similarly, conventional graph coarsening methods, though reducing graph scale, frequently lose fine-grained structural details. MPCCL addresses these challenges through an innovative multi-scale coarsening strategy, which progressively condenses the graph while prioritizing the merging of key edges based on global node similarity to preserve essential structural information. It further introduces a one-to-many contrastive learning paradigm, integrating node embeddings with augmented graph views and cluster centroids to enhance feature diversity, while mitigating feature masking issues caused by the accumulation of high-frequency node weights during multi-scale coarsening. By incorporating a graph reconstruction loss and KL divergence into its self-supervised learning framework, MPCCL ensures cross-scale consistency of node representations. Experimental evaluations reveal that MPCCL achieves a significant improvement in clustering performance, including a remarkable 15.24% increase in NMI on the ACM dataset and notable robust gains on smaller-scale datasets such as Citeseer, Cora and DBLP. In the large-scale Reuters dataset, it significantly improved by 17.84%, further validating its advantage in enhancing clustering performance and robustness. These results highlight MPCCL's potential for application in diverse graph clustering tasks, ranging from social network analysis to bioinformatics and knowledge graph-based data mining. **The source code for this study is available at https://github.com/YF-W/MPCCL**

**Keywords:** Self-supervised Learning; Multi-scale; Attribute Graph Clustering; Graph Coarsening; Contrastive Learning

## 1. Introduction

Since the advent of the information age, the intricate connections between data harbor abundant latent information, from which we can extract a wealth of valuable insights. Attributed graphs build upon traditional graph topology by incorporating node attributes, thereby enhancing the expressiveness of graph features. For example, in natural language processing (NLP), graph structures can be applied to fields like semantic networks and knowledge graphs, where lexical types, entity relationships, and syntactic structures serve as node attributes [1]. Attributed graph clustering methods analyze data by combining topology and node attributes, grouping nodes based on structural and attribute information [2]. This clustering approach can treat words, sentences, and documents as nodes, leveraging topological and attribute information to construct knowledge graphs, and perform document and sentence clustering [3]. Moreover, self-supervised learning techniques have shown significant advantages in attributed graph clustering by generating supervisory signals from the intrinsic structure of the data, thereby reducing dependence on labeled data and producing more expressive feature representations by integrating structural and attribute information [4]. Overall, attributed graph clustering, as a technique that organically combines graph structure with node attributes, not only significantly enhances clustering quality but also emerges as an indispensable tool in data analysis. This technology provides more precise and insightful means for deep data mining, information aggregation, and knowledge discovery, demonstrating irreplaceable value across various research and application domains.

Graph Neural Networks (GNN) have become essential tools for analyzing and processing graph-structured data [5]. Early Recurrent Graph Neural Networks (R-GNN) captured dependencies between nodes through recursive propagation; however, due to high computational complexity and difficulties in parallelization, R-GNN face challenges when applied to large-scale datasets [6]. To improve computational efficiency, Kipf & Welling introduced the graph convolution operation, simplifying the feature aggregation process and achieving outstanding performance in node classification tasks. Nonetheless, shallow Graph Convolutional Networks (GCN) are limited to capturing local information and cannot effectively learn the global structure and long-range dependencies of complex graphs [7]. To enhance the flexibility of feature aggregation, Graph Attention Networks (GAT) introduced a self-attention mechanism that dynamically assigns importance weights to neighboring nodes, giving them an advantage in handling complex relationships. However, as the number of layers increases, GAT still encounter the over-smoothing problem, restricting their ability to capture complex global relationships [8]. To further improve the expressive power of GNN, the Graph Isomorphism Network (GIN) utilizes an injective aggregation function to enhance the model's capacity to distinguish non-isomorphic graphs [9]. Although GIN introduces a more sophisticated mechanism in the aggregation process, it does not assign different weights to different neighboring nodes, treating all neighboring node features equally during aggregation. This approach exacerbates feature convergence, particularly in complex or heterogeneous graph data, leading to potential information loss and impacting the model's overall performance.

In recent years, contrastive learning, as an emerging approach in unsupervised learning, has demonstrated significant advantages in the field of graph representation learning [10]. By constructing positive and negative sample pairs, it maximizes the similarity of positive samples and minimizes the similarity of negative samples, thereby pulling together the embeddings of similar nodes while pushing apart unrelated nodes, capturing the global information of the graph structure. The NCAGC model (Neighborhood Contrastive Attributed Graph Clustering) proposed by Wang et al. selects the most similar neighbors of a node via the KNN method to optimize node representation on the basis of contrastive learning. However, NCAGC has high computational complexity, and the KNN method's reliance on the local neighborhood of nodes may lead to insufficient intra-class diversity, especially in cases where the intra-class data distribution is complex, potentially causing the model to overfit to local features [11]. Furthermore, simplistic positive and negative sample selection may result in high similarity between positive and negative samples [12]. Additionally, single-view approaches that rely on a single augmentation method or identical data patterns can cause feature collapse, leading to a sharp decline in the model's classification or recognition ability [13].

In attributed graph clustering, methods relying on GCNs or incorporating contrastive learning have achieved notable progress; however, these existing methods often focus solely on low-order attribute information while neglecting high-order structural information [14]. GCN-based models generally suffer from a narrow receptive field, making it challenging to effectively capture features of distant nodes within the same class [15]. Although models integrating the graph attention mechanism alleviate this issue to some extent [16], GNN models based on graph attention or Transformer architectures capture information at a single scale, and capturing higher-order information requires stacking multiple network layers, which may lead to overfitting. To address the limitation of single-scale information, Zhiping Lin et al. introduced graph filtering techniques and multi-scale regularization of smoothed node representations to enhance clustering performance. This model explores high-order neighborhood information and conducts multi-scale analysis of the data, thereby obtaining richer feature representations across different levels for clustering, mitigating issues of information deficiency or local optima [17]. Zijing Liu and Mauricio Barahona proposed a graph clustering method based on multi-scale community detection, where their framework examines clustering results through multi-scale analysis, effectively reducing sensitivity to parameter choices in graph construction [18]. However, purely multi-scale techniques alone may not capture the features of distant nodes within the same class. Dhillon et al. proposed a graph coarsening technique that reduces graph size by merging nodes and edges, preserving the original nodes' features and structural information in the newly merged nodes [19]. Nevertheless, multi-level graph coarsening presents the limitation that the coarsened graph cannot be directly applied to self-supervised graph neural clustering networks, as the reduced number of nodes in the coarsened graph makes self-supervised learning of nodes infeasible [20]. Thus, using only the coarsened graph structure without node attributes is a possible solution to this issue.

**Motivations**

Despite significant advancements in attributed graph clustering by integrating graph structure and node attributes, the methods discussed above still face multiple limitations:

**i) GCN-based models struggle to effectively capture long-range dependencies due to their narrow receptive fields.** Most approaches rely excessively on low-order attribute information, overlooking high-order structural information. While the graph attention mechanism has improved this issue to some extent, it can only capture single-scale information, and stacking multiple layers risks overfitting. **ii) In contrastive learning methods, the selection**

**of positive and negative samples is overly simplistic, potentially leading to insufficient intra-class diversity and feature collapse.** Although multi-scale analysis broadens the model's perspective, it struggles to capture the associations between global nodes effectively. **iii) Furthermore, while graph coarsening techniques simplify the graph's scale, they risk losing fine-grained features.** The reduction in the number of nodes after multiple rounds of coarsening limits the effectiveness of self-supervised learning. Given these limitations, this paper proposes a novel approach based on multi-scale graph coarsening and contrastive learning, aiming to better capture both global and local information as well as the hierarchical structure of the graph, thereby enhancing clustering performance.

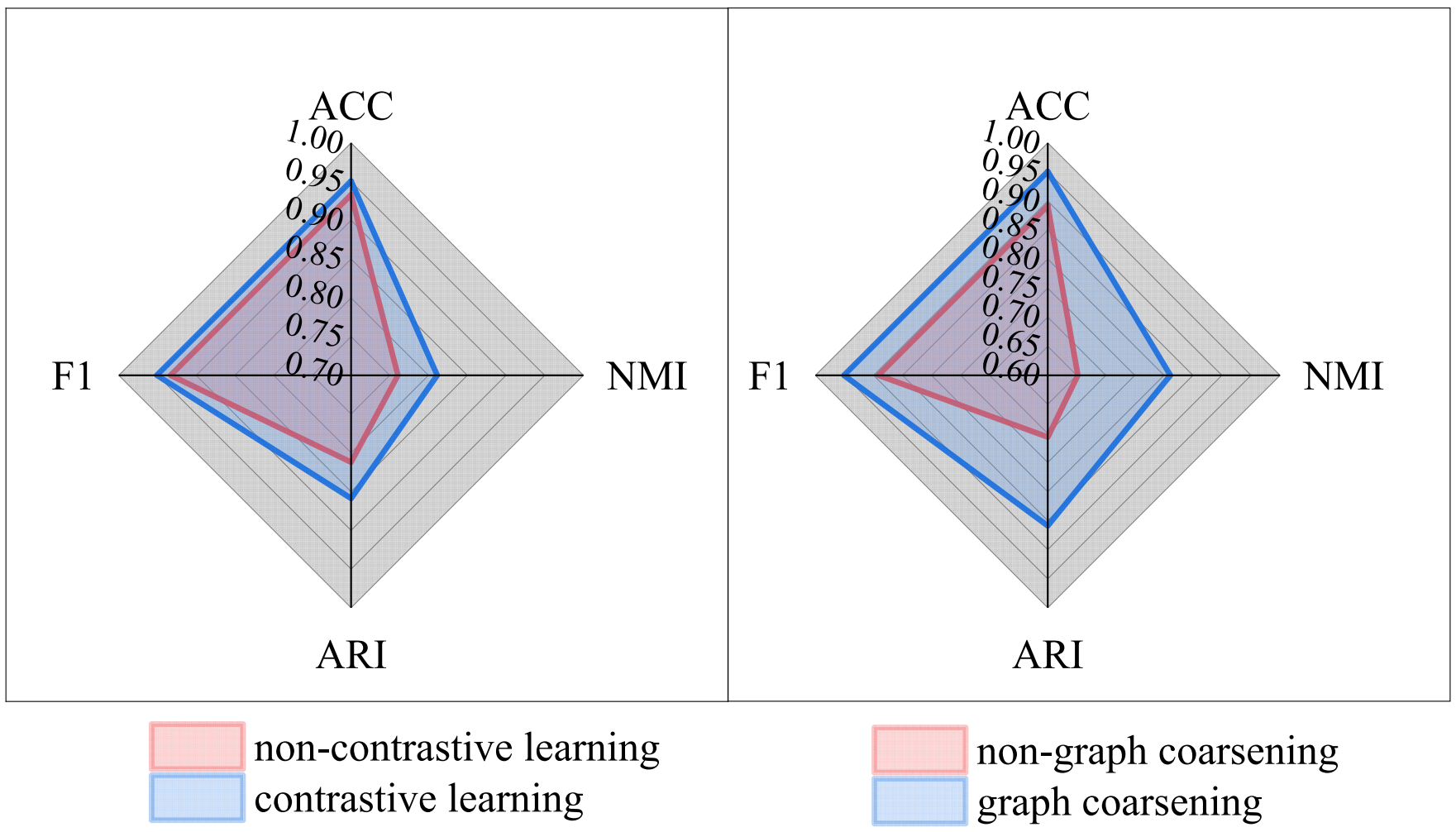

**Fig. 1.** The performance of MPCCL with and without contrastive learning or with and without graph coarsening on four evaluation metrics in the ACM dataset.

We propose a novel attribute graph clustering method based on weight-paired multi-scale graph coarsening and Laplacian regularization contrastive learning, termed MPCCL. Initially, we enhance the data through feature dropout, forcing the model not to rely on specific features during training and providing a second view for contrastive learning. We devise a multi-scale graph coarsening approach to capture structural information from the coarsened graph at different levels. During the coarsening process, we prioritize merging node pairs with higher similarity across the entire graph, rather than solely relying on the similarity of local neighbors. The goal is to ensure that node pairs with global structural significance are preserved, avoiding the loss of global information caused by excessive reliance on local structures. More importantly, when merging, we only retain the structural information between nodes, while the original node features are preserved, preventing the loss of feature information due to excessive coarsening. As shown in Figure 1, on the ACM dataset, the model using multi-scale graph coarsening performs significantly better than the one without it, indicating that capturing multi-level graph structure information through coarsening enhances clustering performance. In contrastive learning, we employ the KMeans clustering method, clustering the node representations from the second view and obtaining cluster centers for each view. By this approach, low-frequency samples within the same cluster are influenced by the contrastive learning of positive pairs, making the features of low-frequency samples in the same cluster more similar to those of high-frequency samples. This enables the preservation of important learning signals during the coarsening process, achieving a complementary mechanism between contrastive learning and multi-scale graph coarsening. Figure 1 compares the models with and without contrastive learning, demonstrating the effectiveness of our contrastive learning design on the ACM dataset. To further enhance model stability and representational capability, we add a Laplacian regularization term in the loss function to ensure smoother representations among neighboring nodes. Additionally, we design a GCN-based encoder to generate a reconstructed adjacency matrix, constructing a reconstruction loss term in the total loss to preserve the structural information of the graph as much as possible.

**Contributions of MPCCL**

**i)** This paper introduces a novel weight-based paired multi-scale graph coarsening method, which prioritizes merging node pairs that exhibit high similarity with the entire graph, thereby avoiding information loss caused by over-reliance on local structures. Simultaneously, during node merging, edge weights are transmitted to ensure that important topological features are preserved throughout the coarsening process. This approach enhances the global nature of the graph structure and the retention of fine-grained information.

**ii)** We have designed a "one-to-many" contrastive learning mechanism with regularization. By comparing the representation of each node with other nodes and their respective cluster centers, we have enhanced the diversity of positive node pairs. This mechanism effectively mitigates the bias towards high-frequency nodes during coarsening, ensuring that the features of low-frequency nodes are also thoroughly learned. Additionally, the introduction of a Laplacian regularization term guarantees the smoothness of node representations over the graph structure, thereby strengthening the model's stability and global consistency.

**iii)** In our network design, we employ a multi-layer GCN as the encoder, enhancing the model’s expressive power by generating adjacency matrix predictions through node reconstruction. An MLP is introduced as the representation projection layer, further improving the resolution of node representations. Our self-supervised learning approach leverages not only the similarity between nodes but also deepens the model’s comprehension of graph structure, thereby enhancing robustness and accuracy attributing graph clustering tasks.

**iv)** Through experimental analysis on five benchmark datasets, we validate MPCCL’s superior performance in clustering real-world graph data. This study's innovations not only advance graph clustering techniques but also introduce new perspectives and practical tools for graph deep learning research, potentially offering fresh insights and applications in fields such as social network analysis and bioinformatics.

**Section Arrangement**

In Section 2, we will briefly review related work in the field of attribute graph clustering, including contrastive learning, multi-scale techniques, and graph coarsening. Section 3 provides a detailed discussion of the design strategies, formulas, and specific implementation details of our proposed method, following the flow of data. Section 4 presents our experiments and result analysis. In Section 5, we discuss the scale sensitivity of the multi-scale approach and specifically analyze the model's performance on the Cora dataset. Finally, Section 6 concludes the paper.

## 2. Related Work

### Graph Clustering Methods Based on GNN

As a key tool for handling non-Euclidean data, GNN captures both the topological structure and attribute information of nodes. By recursively aggregating information from nodes and their neighbors, GNN excel in classification and prediction tasks on graph-structured data. However, GNN primarily rely on aggregating local neighborhood information, which limits their performance in handling long-range dependencies. Additionally, in multi-layer networks, nodes repeatedly aggregate and average their neighbors' features, causing the feature vectors of different nodes to converge, thereby diminishing the model's discriminative power. These issues constrain the application of GNN in complex clustering tasks. To address these limitations, Cross-attention fusion based enhanced graph convolutional network (CaEGCN) enhances the capability to handle long-range dependencies in graph data by integrating edge features and employs a flexible information fusion mechanism to mitigate the over-smoothing problem, achieving impressive performance in clustering tasks on high-dimensional and complex graph data [21]. However, the feature extraction approach of CaEGCN still relies on a static graph structure, which hinders its ability to capture the global relationships within graph data. These limitations have driven the development of deep clustering models based on Transformers. The Transformer model, introduced by Vaswani et al. in 2017 [22], utilizes a self-attention mechanism, enabling parallel processing of input sequences and effectively capturing global relationships. Transformers provide a powerful tool for clustering tasks, particularly for deep clustering. For instance, TDEC (Transformer-based Deep Embedded Clustering) combines deep embedding methods with the Transformer architecture, performing refined clustering of complex data in the embedding space. It leverages the multi-head self-attention mechanism of Transformers to effectively capture global relationships, making it especially suitable for clustering unstructured data [23]. VATC (Variational Autoencoder with Transformer for Clustering) [24] combines the Variational Autoencoder (VAE) [25] with the Transformer architecture, where the former generates latent representations of the data and the latter captures contextual information through the self-attention mechanism, ultimately achieving clustering in the embedding space.

### Multi-Scale Graph Clustering Methods

The core idea of multi-scale techniques is to address the challenges posed by complex structures or hierarchical features by simultaneously processing information at different scales [26]. Early studies introduced multi-scale wavelet analysis, which enabled the discovery and analysis of community structures within graphs across multiple levels, thus achieving scale-based node clustering [27]. In recent years, multi-scale approaches in graph clustering have diversified. Liu et al. proposed a multi-scale community detection method that integrates graph creation with the Markov stability framework, capturing structures from local to global through varying time scales without requiring pre-set cluster numbers [18]. Lipov et al. combined pixel relationships across scales with graph convolution operations, enhancing the model's ability to capture non-Euclidean distances, allowing it to accurately identify local features in images while handling complex structural information across scales [28]. As an integral part of multi-scale techniques, graph coarsening plays a significant role in graph clustering. It progressively reduces the graph's size while retaining key structural information, ensuring effective final clustering. In the pioneering Multilevel Graph

Partitioning Algorithm introduced by Karypis et al., during the coarsening phase, adjacent vertices are collapsed into "multi-nodes," gradually shrinking the graph's size while preserving its core structure, thereby ensuring the quality of the final partition [29]. In recent research, Sun proposed an efficient graph clustering algorithm based on spectral coarsening, which identifies subsets in the original network that are likely to belong to the same cluster and merges them into single nodes. The network's size decreases with coarsening, and spectral clustering is then performed on the coarsened network [30]. Cai et al. parameterized weight allocation in graphs using graph neural networks and trained them in an unsupervised manner to improve coarsening quality [31].

**Graph Clustering Methods Based on Contrastive Learning**

In recent years, contrastive learning has demonstrated remarkable potential in the field of graph clustering [32]. Li et al. first proposed Contrastive Clustering (CC), which performs clustering through instance-level and cluster-level contrastive learning, leveraging data augmentation to optimize representation learning and clustering assignments [33]. Building on CC, Zhu et al. combined contrastive learning with graph clustering, forming Graph Contrastive Learning (GCL) [34]. Most GCL-based clustering methods typically consider corresponding points in two augmented views as positive samples, while other points serve as negative samples. Among these methods, Zhong et al. introduced Graph contrastive clustering (GCC), which elevates instance-level consistency to cluster-level consistency, reducing intra-cluster variance and enhancing inter-cluster discriminability [35]. Zhao et al. proposed Graph Debiased Contrastive Learning with Joint Representation Clustering (GDCL), which combines representation learning and clustering by constructing positive and negative sample pairs with a de-biasing strategy, minimizing the impact of false negative samples and thereby improving both clustering performance and representation learning capability [36]. Wang et al. developed Multi-Graph Contrastive Learning Clustering Network (MGCLCN), which employs a multi-graph contrastive learning mechanism, constructing positive samples from multiple perspectives to optimize node representation [37]. This approach enhances the model's clustering performance and efficiently utilizes graph structural information through clustering methods.

**Limitations**

While GNN have found extensive applications in processing graph data, they rely on local neighborhood information, resulting in limitations when capturing long-range dependencies. Furthermore, as the network depth increases, the convergence of node features weakens the model's discriminative power. Traditional clustering algorithms, such as the Firefly Forest algorithm [38], leverage swarm intelligence for optimization but often struggle to handle high-dimensional data and complex graph structures, highlighting their limitations in dynamic and large-scale graph scenarios. In contrast, multi-scale and graph coarsening techniques offer new approaches for handling graph structural information. However, most existing multi-scale and graph coarsening methods primarily operate at a single level, neglecting the relationships between different scales. Additionally, they typically depend on static graph structures, making them less adaptable to the dynamic changes in complex graph data. Notably, current graph coarsening techniques often fail to effectively retain crucial local features while preserving global structural information, leading to information loss and reduced accuracy. On the other hand, contrastive learning-based clustering methods, though demonstrating significant potential in unsupervised learning, lack specificity in the selection of positive and negative samples. Most graph contrastive learning models use static neighborhood relations to select positive samples, overlooking dynamic changes in graph structures. Moreover, they often fail to fully integrate the semantic information of clustering objectives, limiting the model's performance in capturing complex graph structures. Existing methods have yet to effectively enhance model robustness and generalization in handling contrastive learning tasks on complex graph structures. These limitations indicate research potential in addressing the dynamic changes in graph structures and enhancing the diversity of positive and negative sample selection.

## 3. Methodology

### 3.1 Notation

Let the input data tensor be $x$, and the input tensor after data augmentation be $x_{\text{aug}}$. Feature dropout operation utilizes a binary mask $M \in \{0,1\}^d$ with dropout probability $p$, where $d$ represents the number of features. In the graph $G = (V, E, W)$, $V = \{v_i\}_{i=1}^{m}$ is the set of nodes, $E = \{e_{ij}\}$ is the set of edges, and $W$ is the set of edge weights. The node feature matrix is denoted as $X$, the adjacency matrix as $A$, the degree matrix as $D$, and the symmetrically normalized adjacency matrix as $\hat{A} = D^{-\frac{1}{2}}AD^{-\frac{1}{2}}$. The Laplacian matrix is given by $L = D - A$. The encoder employs a two-layer GCN with weight matrices $W(1)$ and $W(2)$, and an activation function $\sigma$ (e.g., ReLU). The projection head uses an MLP with parameters including weight matrices $W^{(p1)}$, $W^{(p2)}$, and bias terms $b^{(p1)}$ and $b^{(p2)}$.

### 3.2 Overall framework of MPCCL

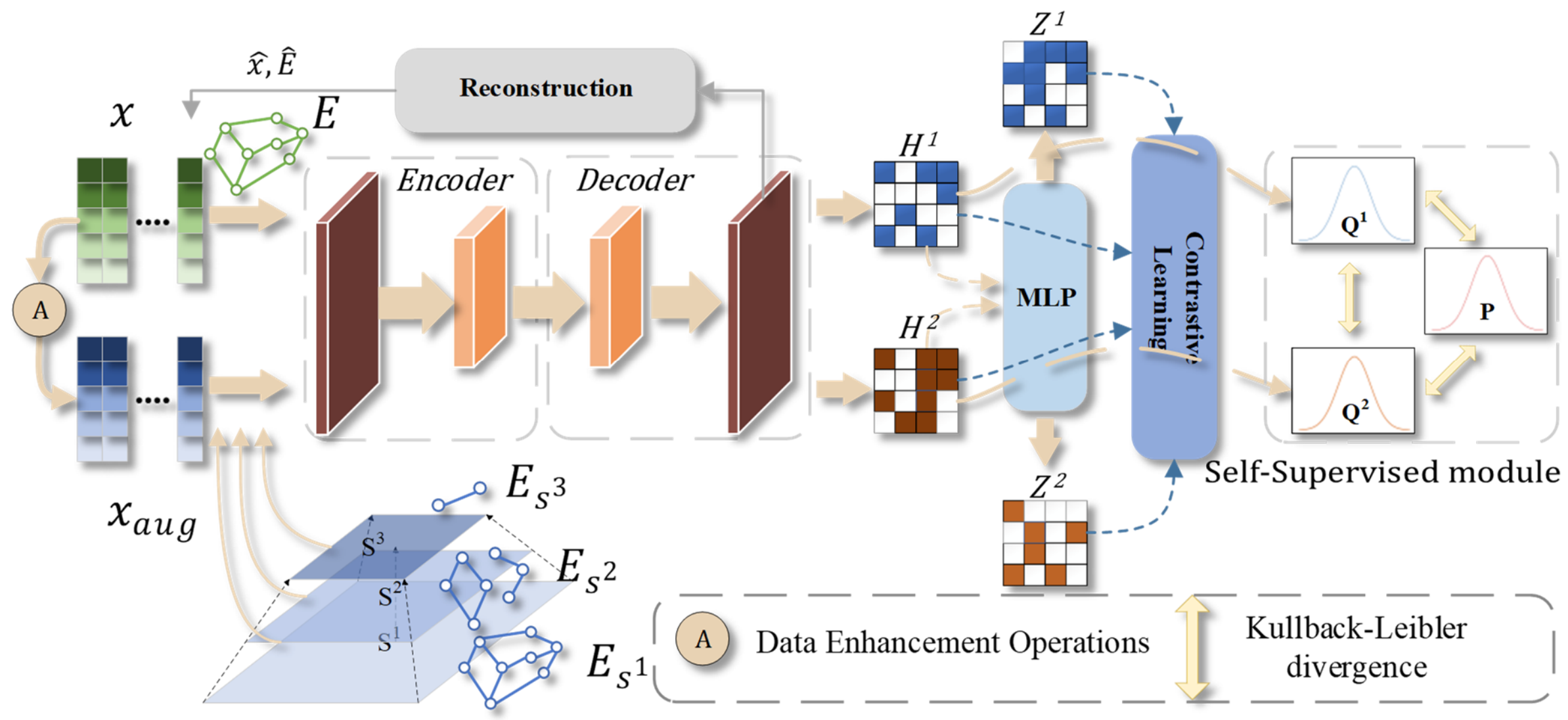

**Fig. 2.** The Overall Framework of MPCCL

The core idea of MPCCL is to optimize node embedding representations through multi-scale graph coarsening and contrastive learning mechanisms, enabling efficient self-supervised clustering. This method constructs a multi-level graph clustering model through four main components: data augmentation, multi-scale graph coarsening, contrastive learning, and self-supervised modules, as illustrated in Figure 2.

### 3.3 Data augmentation

Feature dropout is an effective augmentation strategy that randomly sets a portion of the input features to zero. In our implementation, we designed a data augmentation function responsible for applying feature dropout to the input data. Given an input tensor $x$ and a dropout probability $p$, this function generates a random mask to randomly select features for each input sample. Features corresponding to the mask are set to zero, effectively removing them from the model's input for that iteration. This operation can be mathematically represented as:

$$x_{\text{aug}} = x \cdot (1 - M) \tag{1}$$

where $M \in \{0,1\}^d$ is a binary mask generated based on the dropout probability $p$, with $d$ representing the number of features. The mask is sampled from a uniform distribution, with each feature having a probability $p$ of being set to zero:

$$M_i = \begin{cases} 1 & \text{if } \rho < p, \\ 0 & \text{otherwise,} \end{cases} \tag{2}$$

where $\rho$ is a random value sampled from a uniform distribution $U(0,1)$. This ensures that each feature has an independent dropout probability.

### 3.4 Pairwise Multi-Scale Graph Coarsening

Multi-scale graph coarsening involves progressively reducing the graph at different scales to better capture its multi-layered features, allowing the model to learn a comprehensive representation of both global and local information across various scales. To enable effective node matching and merging, it is necessary to define the similarity between nodes. We use the cosine similarity of node features to calculate edge weights. For nodes $u$ and $v$, with feature vectors $\mathbf{x}_u$ and $\mathbf{x}_v$, the edge weight $w(u, v)$ is calculated as follows:

$$w(u, v) = \cos(\mathbf{x}_u, \mathbf{x}_v) = \frac{\mathbf{x}_u^\top \mathbf{x}_v}{\| \mathbf{x}_u \| \| \mathbf{x}_v \|} \tag{3}$$

We assign the similarity of node features to the corresponding edges, resulting in a weighted graph $G = (V, E, W)$, where $W$ is the set of edge weights. This approach gives edges between similar nodes higher connection weights, emphasizing these similarities within the graph structure. This weighted graph provides a foundation for our pairwise node matching theory. These edge weights then serve as the foundation for the subsequent pairwise node matching theory, wherein maximum similarity matching is employed.

**Pairwise Node Matching Theory**

The goal of our proposed pairwise node matching is to pair and merge similar nodes based on their similarity, reducing the graph's size. To theoretically describe this process, we define a matching relation and a merging operation. Define a matching function $M: V \rightarrow V$, satisfying the following conditions:

**Symmetry**. For any $u \in V$, if $M(u) = v$ and $u \neq v$, then $M(v) = u$.

**Disjointness**. For any $M(u) = v$, if $M(u) = v$, then $M(v) = u$ and $u \neq v$.

**Maximum Similarity Matching**. For each unmatched node $u$ and $v$, the matching function is defined as $M(u) = \arg max\, w\,(u, v)$.

Given the edge sets $E(u) = \{x|(u, x) \in \mathrm{E}\}$ and $E(v) = \{x|(v, x) \in \mathrm{E}\}$, the merged node set is formulated as $C(u, v) = \big(E(u) \cup E(v)\big)\{u, v\}$. Here, $E(u)$ represents the set of nodes that include $u$ itself and all nodes connected to it, while $E(v)$ represents the set of nodes that include $v$ itself and all nodes connected to it. The set $C(u, v)$ denotes the collection of nodes obtained after merging $u$ and $v$, along with their connected neighbors, as illustrated in Figure 3.

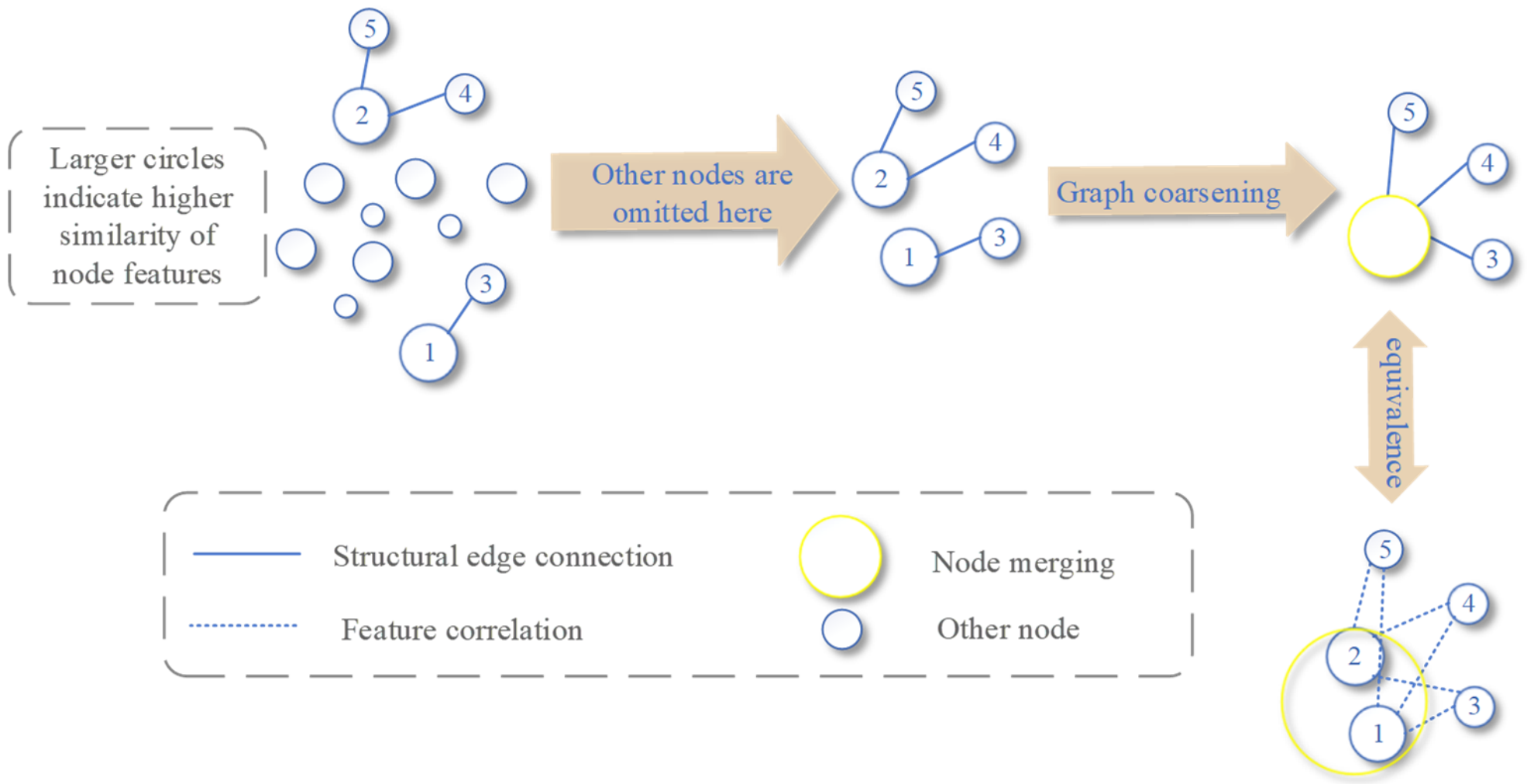

**Fig. 3.** Schematic Diagram of Node Merging in Graph Coarsening

The matching function $M$ partitions the nodes of the graph into matching pairs $(u, v)$, ensuring that each node belongs to at most one pair. For each matching pair $(u, v)$, we define a node merging operation to create a new node $n$. Since we only need multi-scale structural information and not feature vectors for the new node, we update the edge weights as follows: $w(n, w) = w(u, w) + w(v, w)$. Unconnected edge weights are considered zero. Based on the above theory, each merging operation combines two nodes into one, reducing the total number of nodes in the graph. By accumulating edge weights, the new edge weight reflects the combined node's overall association with other nodes.

**Multi-Scale Graph Coarsening Process**

To capture the multi-level structural information of the graph, we perform coarsening at various scales. Given a set of scale parameters $s_1, s_2, \ldots, s_K$, each corresponds to a different graph scaling ratio. For each scale $s_k$, we first determine the target number of nodes: $N_k = max(N_{min}, \lfloor s_k N \rfloor)$, where $N_{min}$ is a predefined minimum node count, and $N$ is the node count of the original graph. We then proceed to the matching stage, where we generate a set of matching pairs $M$ by using the matching function $\mathcal{M}$ based on the edge weights of the current graph. Specifically, the matching function $\mathcal{M}$ is an algorithm or procedure that selects pairs of nodes to be merged based on edge weights, structural similarity, or connectivity, while the set $M$ refers explicitly to the resulting collection of node pairs identified by this matching function. After matching is complete, in the merging stage, we perform a node merging operation for each matching pair $(u, v) \in \mathcal{M}$, updating the node features and edge set. The matching and merging steps are repeated until the node count of the graph is reduced to the target $N_k$. By repeating this process across different scales, we obtain a series of coarsened graphs $G^{(s)} = \{G^{(s_1)}, G^{(s_2)}, \dots, G^{(s_K)}\}$, where each coarsened graph captures the structural information of the original graph at its respective scale. These multi-scale coarsened graphs can serve as inputs to the network, as coarsening reduces noise and the complexity of local structures. By clustering at different coarsening scales, the model can capture clustering features across scales, thus enhancing the effectiveness of node representation learning.

**Assumption 1.** Partition the set of nodes $V$ into $n'$ disjoint subsets $C_1, C_2, \dots, C_{n'}$, such that $V = \bigsqcup_{i=1}^{n'} C_i$. For any $C_i$ and $C_j$ (where $i \neq j$), the edge weights between any nodes $u, u' \in C_i$ and $v \in C_j$ are identical, meaning there exists a constant $k_{ij}$ such that $w_{uv} = w_{u'v} = k_{ij}, \forall u, u' \in C_i, \forall v \in C_j$. In this definition, edge weights within the same subset are equal, and the edge weights between different subsets are represented by the constant $k_{ij}$.

**Assumption 2.** By merging the nodes within each subset $C_i$ into a new node $n_i$, we construct the coarsened graph $G^{(s)}$. The Laplacian matrix $L'$ of the coarsened graph is defined as:

$$L' = P^\top L P, \tag{4}$$

where $L$ is the Laplacian matrix of the original graph $G$, and $P$ is the projection matrix defined in Assumption 3.

**Assumption 3.** When constructing the projection matrix $P$, assume that each subset $C_i$ contains $|C_i|$ nodes, and uniformly distributes the node feature vector. Specifically, define:

$$P_{un_i} = \begin{cases} \dfrac{1}{\sqrt{|C_i|}}, & \text{if node } u \in C_i, \\ 0, & \text{otherwise.} \end{cases} \tag{5}$$

Consequently, the projection matrix $P$ satisfies $P^\top P = I_{n'}$, where $I_{n'}$ is the $n' \times n'$ identity matrix.

**Theorem 1.** If **Assumptions 1**, **2**, and **3** hold, let $G = (V, E)$ be an undirected graph with Laplacian matrix $L \in \mathbb{R}^{n \times n}$. By merging nodes, $G$ is coarsened into the graph $G' = (V', E')$, with Laplacian matrix $L' \in \mathbb{R}^{n' \times n'}$, where $n' < n$ and $L' = P^\top L P$, Then the eigenvalues of $L$ and $L'$ satisfy:

$$\lambda_k(L) \leq \lambda_k(L') \text{ for } k = 1, 2, \dots, n', \tag{6}$$

where $\lambda_k(L)$ and $\lambda_k(L')$ denote the $k$-th eigenvalues of $L$ and $L'$, respectively, arranged in non-decreasing order.

**Remark 1.** Graph coarsening reduces the graph's scale by merging similar nodes. In this process, each new node generally represents a cluster of original nodes, and its connection weight with other nodes is the cumulative sum of the original connection weights. Since the merging operation is typically based on high similarity (i.e., strong connectivity) between nodes, the coarsened graph exhibits stronger connectivity between these merged nodes. This enhanced connectivity is reflected in the eigenvalues of the Laplacian matrix, particularly in the increased algebraic connectivity. The increase in Laplacian eigenvalues indicates stronger graph connectivity, suggesting that the coarsened graph is, in a sense, more "compact." This compactness aids the network part of the model in learning more effective node representations, enhancing the discriminative power of the representations.

**Assumption 4.** The graphs $G$ and its coarsened graph $G^{(s)}$ are connected undirected graphs. Consequently, their Laplacian matrices $L$ and $L^{(s)}$ are symmetric positive semidefinite matrices with a single zero eigenvalue corresponding to the constant vector.

**Lemma 1.** For a symmetric positive semidefinite matrix $A \in \mathbb{R}^{n \times n}$ with eigenvalues $0 = \lambda_1(A) \leq \lambda_2(A) \leq \cdots \leq \lambda_n(A)$, the condition number is defined as:

$$\kappa(A) = \frac{\lambda_{max}(A)}{\lambda_{min}^{+}(A)}, \tag{7}$$

where $\lambda_{min}^{+}(A) = \lambda_2(A)$ is the smallest positive eigenvalue (also known as the algebraic connectivity), and $\lambda_{max}(A) = \lambda_n(A)$ is the largest eigenvalue of $A$.

**Theorem 2-1.** Under Assumption 4 and using Lemma 1, the condition number of the coarsened Laplacian matrix $L^{(s)}$ is no greater than that of the original Laplacian matrix $L$:

$$\kappa\left(L^{(s)}\right) \leq \kappa(L). \tag{8}$$

**Theorem 2-2.** There exists a projection matrix $P \in \mathbb{R}^{n \times n^{(s)}}$ such that

$$\| L - PL^{(s)}P^{\top} \|_2 \leq \epsilon, \tag{9}$$

where $\epsilon$ is a small constant, indicating that $L^{(s)}$ is a good spectral approximation of $L$.

**Remark 2.** A smaller condition number implies reduced error propagation and increased stability in numerical computations involving the Laplacian matrix (such as solving linear systems, eigenvalue decomposition, etc.). When computing $y = L^{-1}x$ with a small error $\delta x$ in the input vector $x$, the output error $\delta y$ satisfies:

$$\frac{\| \delta y \|}{\| y \|} \leq \kappa(L) \cdot \frac{\| \delta x \|}{\| x \|}. \tag{10}$$

Since $\kappa(L^{(s)}) \leq \kappa(L)$, the coarsened Laplacian matrix is more stable in numerical computations, with less error amplification. Additionally, in training graph neural networks, a smaller condition number of aids optimization algorithms like gradient descent to converge faster, improving training efficiency.

**Remark 3. Theorem 2-2** presents a significant projection approximation result. This indicates that the coarsened graph not only has an advantage in terms of condition number but also can accurately approximate the spectral properties of the original graph through the projection matrix $P$. In other words, the coarsened graph, in graph neural network training, can retain the structural information of the original graph via precise spectral approximation. This is particularly beneficial for GCN training, as it ensures that while using the coarsened graph, the spectral characteristics of the Laplacian matrix are preserved, enhancing the stability and performance of the training process.

In this module, although our model's multi-scale graph coarsening method effectively addresses many challenges of traditional graph coarsening, certain issues may still persist. High-frequency nodes, which are connected to a larger number of other nodes and tend to exhibit greater similarity among themselves, may lead the model to continuously merge the edge weights between similar nodes during the coarsening process. Consequently, this can result in the GCN propagation process excessively focusing on feature updates for high-frequency samples while overlooking the learning of low-frequency nodes. To address this issue, we have designed a "one-to-many" contrastive learning mechanism within our model, which effectively alleviates this problem.

### 3.5 Network

In graph data, the role and function of nodes are determined by both their features and topological structure. However, traditional graph representation learning methods often capture only a fraction of this information, making it challenging to fully exploit both global and local properties of the graph. Additionally, as the graph scale increases, computational efficiency and storage overhead become pressing issues. We preprocessed the original graph using multi-scale graph coarsening and employ a collaborative encoder-decoder network structure for node representation learning.

We use a two-layer GCN as the encoder to encode the node features. Given the input features $X$ and adjacency matrix $A$, the node embedding $H$ is calculated as follows:

$$H^{(1)} = \sigma\left(\hat{A}XW^{(1)}\right), \tag{11}$$

$$H = \hat{A}H^{(1)}W^{(2)}, \tag{12}$$

Single-scale node representation learning may not adequately capture subtle differences between nodes. Therefore, we enhance the discriminative power of node representations through contrastive learning on different input views (original and augmented graphs). For contrastive learning, we introduce a MLP as a projection head to map the node embeddings $H$ into the contrastive space:

$$Z = \mathrm{MLP}(H) = W^{(p2)}\phi\left(HW^{(p1)} + b^{(p1)}\right) + b^{(p2)}, \tag{13}$$

where $\phi$ is the activation function (we use PReLU).

We perform two types of augmentations on the input data. The first is randomly dropping some features of nodes with a certain probability, yielding an augmented feature matrix $X^{(aug)}$. The second is using coarsened graphs of different scales $G^{(s)}$, resulting in varied graph structures. For the original and augmented graphs, we compute the node embeddings separately. For the original view:

$$\begin{aligned} H^{(1)} &= \text{Encoder}(X, A), \\ Z^{(1)} &= \text{MLP}(H^{(1)}). \end{aligned} \tag{14}$$

For the augmented views (for each scale $s$):

$$\begin{aligned} H^{(2,s)} &= \text{Encoder}(X^{(aug)}, A^{(s)}), \\ Z^{(2,s)} &= \text{MLP}(H^{(2,s)}). \end{aligned} \tag{15}$$

To further enhance the performance of node representation learning, we propose a multi-scale feature weighted averaging fusion strategy to integrate graph features across different scales. After multi-scale graph coarsening, the graph features from various scales are weighted and fused to ensure that information from each scale contributes to the final node representation. We perform weighted averaging on the multi-scale embeddings:

$$H^{(2)} = \frac{1}{K}\sum_{s=1}^{K} w_s H^{(2,s)}, \tag{16}$$

$$Z^{(2)} = \frac{1}{K}\sum_{s=1}^{K} w_s Z^{(2,s)}, \tag{17}$$

where $K$ represents the number of scales, $w_s$ is the weight for scale $s$, and $Z^{(2,s)}$ is the node feature matrix at scale $s$. Through this design, our model can effectively learn robust and discriminative node representations when processing graph data that has undergone multi-scale graph coarsening. The learned representations are subsequently used in the contrastive learning module.

### 3.6 One-to-Many Contrastive Learning

The contrastive learning module we designed enhances node embedding representations by contrasting feature representations, effectively capturing information within the graph structure to achieve efficient graph clustering. The core concept involves introducing cluster centroids as positive pairs within the contrastive loss, ensuring that node embeddings not only resemble their own augmented views but also align closely with the centroids of their respective clusters. This effectively enhances the clustering structure of embeddings, enabling the model to leverage positive contrastive learning to influence low-frequency samples (i.e., nodes with low degrees) that belong to the same cluster as high-frequency samples (i.e., nodes with high degrees). Consequently, low-frequency samples become more similar to high-frequency ones, effectively addressing potential shortcomings in multi-scale graph coarsening methods and aiding the model in learning more discriminative node representations. As shown in Figure 4, the structure of the one-to-many contrastive learning module is illustrated. By constructing positive and negative pairs for feature representation contrast, with the cluster centroids serving as additional positive samples, this module strengthens the clustering structure of node embeddings, leading to more distinct node representations. The entire process includes similarity calculation, positive and negative pair construction, contrastive loss computation, and Laplacian regularization, ultimately optimizing node embeddings for graph clustering tasks.

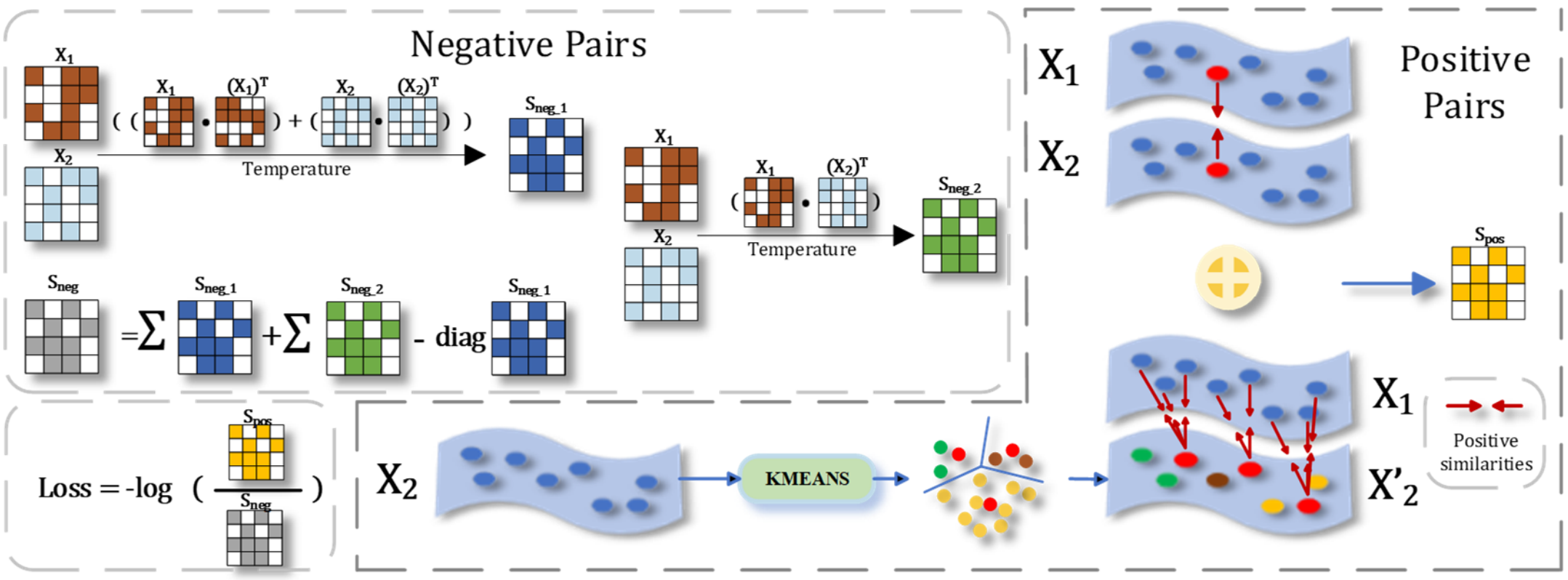

**Fig. 4.** Structure of the Contrastive Learning Module

Given the node representations $\mathbf{z}_1$ and $\mathbf{z}_2$ of two views, we first apply $L_2$ normalization: $\tilde{\mathbf{z}}_1 = \frac{\mathbf{z}_1}{\|\mathbf{z}_1\|_2}, \tilde{\mathbf{z}}_2 = \frac{\mathbf{z}_2}{\|\mathbf{z}_2\|_2}$.

Introducing a temperature parameter $\tau > 0$, the similarity function is defined as: $s_{ij}^{(ab)} = \exp\left(\frac{\tilde{z}_i^{(a)\top}\tilde{z}_j^{(b)}}{\tau}\right)$, where $a, b \in \{1,2\}$.

**Positive Sample Pair Construction**

For each node $i$, its representations $\mathbf{z}_{1i}$ and $\mathbf{z}_{2i}$ in the two views form a positive sample pair. This pair is based on the representation of the same node in different views, guiding the model to learn consistent node representations. The cross-view positive pair similarity is defined as: $s_{ii}^{(12)} = \exp\left(\frac{\tilde{\mathbf{z}}_1^{(i)\top}\tilde{\mathbf{z}}_2^{(i)}}{\tau}\right)$. To incorporate more positive samples in contrastive learning, we perform clustering on the embeddings and use the cluster centroids as additional positive samples. Applying KMeans clustering on $\tilde{\mathbf{z}}_2$, we obtain $M$ cluster centroids $\{\boldsymbol{\mu}_1, \boldsymbol{\mu}_2, \ldots, \boldsymbol{\mu}_M\}$ and corresponding cluster labels $\{c_i\}_{i=1}^N$, where $c_i \in \{1,2,\ldots,M\}$ indicates the cluster to which node $i$ belongs.

The objective function of KMeans clustering is:

$$\min_{\{\boldsymbol{\mu}_j\},\{c_i\}} \sum_{i=1}^{N} \left\|\tilde{\mathbf{z}}_i^{(2)} - \boldsymbol{\mu}_{c_i}\right\|_2^2 \tag{18}$$

The clustering assignment step is:

$$c_i = \arg\min_j \left\|\tilde{\mathbf{z}}_i^{(2)} - \boldsymbol{\mu}_j\right\|_2^2. \tag{19}$$

The update step for the cluster centroids is:

$$\boldsymbol{\mu}_j = \frac{1}{N_j} \sum_{i:c_i=j} \tilde{\mathbf{z}}_i^{(2)}, \tag{20}$$

where $N_j$ is the number of samples in cluster $j$. The similarity for the cluster centroid positive pair is defined as: $s_i^{(1c)} = \exp\left(\frac{\tilde{\mathbf{z}}_1^{(i)\top}\boldsymbol{\mu}_{c_i}}{\tau}\right)$. By introducing cluster centroids as positive pairs in contrastive learning, the model can enhance the discriminative power of embeddings, as node embeddings are required to be similar not only to their own augmented views but also to the centroids of their respective clusters. This strengthens the explicitness of the clustering structure while maintaining the efficiency of contrastive learning.

Positive Sample Total Similarity:

$$s_i^{\text{pos}} = s_{ii}^{(12)} + s_i^{(1c)} \tag{21}$$

For each node $i$, we select representations with other nodes $j \neq i$ to form negative sample pairs. Specifically, $i$'s representations $\tilde{\mathbf{z}}_1^{(i)}$ and $\tilde{\mathbf{z}}_2^{(i)}$ in views 1 and 2, respectively, form negative sample pairs with node $j$'s representations $\tilde{\mathbf{z}}_1^{(j)}$ and $\tilde{\mathbf{z}}_2^{(j)}$. These negative pairs are used to train the model to distinguish between different node representations. The total similarity of negative samples (including intra-modal and cross-modal) is calculated as:

$$s_i^{\text{neg}} = \sum_{j=1}^{N} \left(s_{ij}^{(11)} + s_{ij}^{(12)}\right) - s_{ii}^{(11)} + \epsilon, \tag{22}$$

where $\epsilon$ is a smoothing term to avoid division by zero.

**Contrastive Loss Calculation**

For each node $i$, the contrastive loss is defined as:

$$\mathcal{L}_{contrast} = \frac{1}{N}\sum_{i=1}^{N} -log\left(\frac{s_{ii}^{(12)} + s_i^{(1c)}}{\sum_{j=1}^{N}\left(s_{ij}^{(11)} + s_{ij}^{(12)}\right) - s_{ii}^{(11)} + \epsilon}\right) \tag{23}$$

For the other view $\tilde{\mathbf{z}}_2$, similarly calculate $\mathcal{L}'_{\text{contrast}}$:

$$\mathcal{L}'_{\text{contrast}} = \frac{1}{N}\sum_{i=1}^{N} -\log\left(\frac{s_{ii}^{(21)} + s_i^{(2c)}}{\sum_{j=1}^{N}\left(s_{ij}^{(22)} + s_{ij}^{(21)}\right) - s_{ii}^{(22)} + \epsilon}\right) \tag{24}$$

The total contrastive loss is:

$$\mathcal{L}_{\text{cont}} = \mathcal{L}_{\text{contrast}} + \mathcal{L}'_{\text{contrast}} \tag{25}$$

To preserve the global structure, a Laplacian regularization term is added. First, compute the intra-modal similarity matrix: $\mathbf{S}^{(a)} = \left[s_{ij}^{(aa)}\right]_{i,j=1}^{N}, a \in \{1,2\}$. Compute the degree matrix: $\mathbf{D}^{(a)} = \text{diag}\left(\sum_{j=1}^{N} s_{ij}^{(aa)}\right)$. Calculate the unnormalized Laplacian matrix: $\mathbf{L}^{(\mathrm{a})} = \mathbf{D}^{(\mathrm{a})} - \mathbf{S}^{(\mathrm{a})}$. The normalized Laplacian matrix is: $\mathbf{L}_{\mathrm{norm}}^{(\mathrm{a})} = (\mathbf{D}^{(\mathrm{a})})^{-\frac{1}{2}}\mathbf{L}^{(\mathrm{a})}(\mathbf{D}^{(\mathrm{a})})^{-\frac{1}{2}} = \mathbf{I} - (\mathbf{D}^{(\mathrm{a})})^{-\frac{1}{2}}\mathbf{S}^{(\mathrm{a})}(\mathbf{D}^{(\mathrm{a})})^{-\frac{1}{2}}$. The Laplacian regularization term is: $\mathcal{L}_{\mathrm{lap}}^{(a)} = \mathrm{Tr}\left(\tilde{\mathbf{Z}}^{(a)\top}\mathbf{L}_{\mathrm{norm}}^{(a)}\tilde{\mathbf{Z}}^{(a)}\right)$. The total Laplacian regularization loss is: $\mathcal{L}_{\mathrm{lap}} = \frac{1}{2N}\left(\mathcal{L}_{\mathrm{lap}}^{(1)} + \mathcal{L}_{\mathrm{lap}}^{(2)}\right)$. Combining the contrastive loss and Laplacian regularization term, the total contrastive learning loss is:

$$\mathcal{L}_{\mathrm{total}}^{\mathrm{contrast}} = \mathcal{L}_{\mathrm{contrast}} + \lambda_{reg}\mathcal{L}_{\mathrm{reg}} \tag{26}$$

where $\lambda_{\mathrm{reg}}$ is the regularization weight hyperparameter.

### 3.7 Self-Supervised Module

In this model's self-supervised learning approach, the central goal is to refine node embeddings via clustering optimization, while simultaneously preserving the original graph's topology. To achieve these objectives, the model incorporates graph reconstruction loss and clustering loss. The KL divergence loss specifically measures the distributional discrepancy during clustering, aiding the model in maintaining consistency and distinctiveness of node embeddings across different scales.

**Graph Reconstruction Loss**

To ensure that the generated node embeddings effectively reconstruct the graph's structural information, the model introduces a graph reconstruction loss. This loss primarily aims to minimize the difference between the adjacency matrix of the original graph, $A$, and the reconstructed adjacency matrix, $\hat{A}$, which is generated based on the node embeddings $Z$. The graph reconstruction loss is defined as follows:

$$\mathcal{L}_{\mathrm{reconstruction}} = \| A - \hat{A} \|_F^2 \tag{27}$$

where $\hat{A}$ is obtained by computing the adjacency matrix's reconstructed representation using node embeddings, typically expressed as:

$$\hat{A} = \sigma(ZZ^T) \tag{28}$$

Here, $\sigma$ represents the Sigmoid activation function, mapping the matrix values into the range [0,1]. By minimizing the error between $A$ and $\hat{A}$, this loss ensures that the embeddings generated by the model can optimally reconstruct the original graph's structural relationships, thus capturing the graph's global topological information within the node embeddings.

**KL Divergence Loss**

The model obtains both the original representation $H^{(1)}$ and multi-scale representation $H^{(2)}$ of each node, then applies these representations to perform soft clustering assignments, resulting in cluster distributions $q_1$ and $q_2$. By minimizing the following three KL divergence losses, the model refines node embeddings, guiding them toward cluster centroids while ensuring consistency across representations of different scales:

For each node $z_i$ and cluster center $C_j$, the soft assignment probability $q_{ij}$ is calculated via a Student-t distribution, translating the distance between node representations and cluster centroids into a probability. The soft clustering distributions $q_1$ and $q_2$ are defined as follows:

$$q_{1,ij} = \frac{\left(1 + \frac{\|z_{1,i} - \mu_j\|^2}{v}\right)^{-\frac{v+1}{2}}}{\sum_{j'} \left(1 + \frac{\|z_{1,i} - \mu_{j'}\|^2}{v}\right)^{-\frac{v+1}{2}}} \quad (29)$$

$$q_{2,ij} = \frac{\left(1 + \frac{\|z_{2,i} - \mu_j\|^2}{v}\right)^{-\frac{v+1}{2}}}{\sum_{j'} \left(1 + \frac{\| z_{2,i} - \mu_{j'} \|^2}{v}\right)^{-\frac{v+1}{2}}} \quad (30)$$

where $v$ is a degree of freedom parameter controlling the shape of the distribution. The target distribution $p_{ij}$ is defined as:

$$p_{ij} = \frac{\frac{q_{ij}^2}{f_j}}{\sum_{j'} \left(\frac{q_{ij'}^2}{f_{j'}}\right)}, \quad (31)$$

where $f_j = \sum_i q_{ij}$ represents the soft assignment frequency for cluster $j$, capturing the total assignment probability of all nodes to this cluster. $f_{j'}$ normalizes across potential cluster indices.

$\mathcal{L}_{\text{KL}_1}$ measures the divergence between the target distribution $p_1$ and the soft clustering distribution $q_1$. Minimizing $\mathcal{L}_{\text{KL}_1}$ enhances the clustering accuracy and consistency of the original representation:

$$\mathcal{L}_{\text{KL}_1} = \text{KL}(p_1 \parallel q_1) = \sum_i \sum_j p_{ij} \log\left(\frac{p_{ij}}{q_{1,ij}}\right), \quad (32)$$

$\mathcal{L}_{\text{KL}_2}$ quantifies the divergence between the target distribution $p_1$ and the soft clustering distribution $q_2$. By minimizing $\mathcal{L}_{\text{KL}_2}$, the model's robustness and generalization are improved:

$$\mathcal{L}_{\text{KL}_2} = \text{KL}(p_1 \parallel q_2) = \sum_i \sum_j p_{ij} \log\left(\frac{p_{ij}}{q_{2,ij}}\right). \quad (33)$$

$\mathcal{L}_{\text{KL}_{\text{con}}}$ measures the discrepancy between the soft clustering distributions $q_1$ and $q_2$. Minimizing $\mathcal{L}_{\text{KL}_{\text{con}}}$ ensures consistency in clustering results across different scales for the same node, thereby enhancing the model's clustering stability and adaptability to multi-scale data [39]:

$$\mathcal{L}_{\mathrm{KL}_{\mathrm{con}}} = \mathrm{KL}(q_1 \parallel q_2) = \sum_i \sum_j q_{ij} \log\left(\frac{q_{1,ij}}{q_{2,ij}}\right). \tag{34}$$

The clustering loss, $\mathcal{L}_{\mathrm{clu}}$, is iteratively optimized to improve the model's feature learning, using KL divergence to measure the differences between the target distribution $P$ and the soft clustering distribution $Q$, thereby enhancing overall robustness and clustering performance [40]:

$$\mathcal{L}_{\mathrm{clu}} = \mathcal{L}_{\mathrm{KL}_1} + \mathcal{L}_{\mathrm{KL}_2} + \mathcal{L}_{\mathrm{KL}_{\mathrm{con}}}. \tag{35}$$

The total loss function combines contrastive loss, clustering loss, and graph structure reconstruction loss, defined as:

$$\mathcal{L}_{\mathrm{total}} = \alpha \mathcal{L}_{\mathrm{total}}^{\mathrm{contrast}} + \beta \mathcal{L}_{\mathrm{clu}} + \gamma \mathcal{L}_{\mathrm{reconstruction}} \tag{36}$$

where $\mathcal{L}_{\mathrm{contrastive}}$ represents the contrastive learning loss, and $\alpha$, $\beta$, and $\gamma$ are weights assigned to each loss component, all set to 1 in this model. This design enables the model to optimize node embedding similarity, structural integrity, and clustering performance simultaneously.

### 3.8 Optimization

The proposed model utilizes an integrated optimization framework to jointly refine node embeddings, cluster assignments, and graph reconstruction. The total loss function is a weighted sum of the contrastive loss, clustering loss, and graph reconstruction loss, defined as Eq. (36). To optimize the network parameters and cluster centroids $\mu_j$, we compute the gradients of the total loss with respect to these variables. The gradient of the clustering loss $L_{\mathrm{clu}}$ with respect to the cluster centroids $\mu_j$ is derived as follows.

For the soft assignment probabilities $q_{,ij}$ and $q_{2,ij}$ defined in Equations (29) and (30), the gradients with respect to $\mu_j$ are:

$$\frac{\partial q_{k,ij}}{\partial \mu_j} = -\frac{(v+1)\left(1+\frac{\parallel z_{k,i}-\mu_j \parallel^2}{v}\right)^{-\frac{v+3}{2}} \frac{2\left(z_{k,i}-\mu_j\right)}{v}}{\sum_{j'}\left(1+\frac{\parallel z_{k,i}-\mu_{j'} \parallel^2}{v}\right)^{-\frac{v+1}{2}}} \tag{37}$$

where $k = 1,2$ corresponds to the two scales.

The gradient of the KL divergence losses $\mathcal{L}_{\mathrm{KL}_1}$ and $\mathcal{L}_{\mathrm{KL}_2}$ with respect to $\mu_j$ are:

$$\frac{\partial L_{\mathrm{KL}_k}}{\partial \mu_j} = \sum_i \left(\frac{\partial KL(p \parallel q_k)}{\partial q_{k,ij}} \cdot \frac{\partial q_{k,ij}}{\partial \mu_j}\right) \tag{38}$$

Since $KL(p \parallel q_k) = \sum_i \sum_j p_{ij} \log\left(\frac{p_{ij}}{q_{k,ij}}\right)$, we have:

$$\frac{\partial KL(p \parallel q_k)}{\partial q_{k,ij}} = -\frac{p_{ij}}{q_{k,ij}} \tag{39}$$

Combining Equations (37), (38), and (39), the gradient with respect to $\mu_j$ becomes:

$$\frac{\partial L_{\mathrm{KL}_k}}{\partial \mu_j} = -\sum_i \frac{p_{ij}}{q_{k,ij}} \cdot \frac{\partial q_{k,ij}}{\partial \mu_j} \tag{40}$$

Similarly, the gradient of the consistency loss $\mathcal{L}_{\mathrm{KL}_{\mathrm{con}}}$ with respect to $\mu_j$ is:

$$\frac{\partial L_{\mathrm{KL}_{\mathrm{con}}}}{\partial \mu_j} = \sum_i \left( \frac{\partial KL(q_1 \parallel q_2)}{\partial q_{1,ij}} \cdot \frac{\partial q_{1,ij}}{\partial \mu_j} + \frac{\partial KL(q_1 \parallel q_2)}{\partial q_{2,ij}} \cdot \frac{\partial q_{2,ij}}{\partial \mu_j} \right) \tag{41}$$

where: $\frac{\partial KL(q_1 \| q_2)}{\partial q_{1,ij}} = \log\left(\frac{q_{1,ij}}{q_{2,ij}}\right) + 1$ and $\frac{\partial KL(q_1 \| q_2)}{\partial q_{2,ij}} = -\frac{q_{1,ij}}{q_{2,ij}}$.

**Convergence and Output**

Upon convergence, the model yields optimized node embeddings that accurately reconstruct the graph structure and exhibit improved clustering performance. The final soft assignments $q_{1,ij}$ and $q_{2,ij}$ provide the clustering results for each node. The predicted cluster label $r_i$ for node iii can be obtained by:

$$r_i = \arg\max_j q_{ij} \tag{42}$$

where $q_{ij}$ can be either $q_{1,ij}$ or an average of $q_{1,ij}$ and $q_{2,ij}$ to incorporate multi-scale information.

### 3.9 Complexity Analysis

The time complexity of this model primarily arises from the following aspects: the multi-scale graph coarsening process, the forward propagation of the GCN encoder, computations in the contrastive learning module, and the loss calculations in the self-supervised module. Firstly, during the multi-scale graph coarsening, for each scale, we need to compute similarities between nodes, perform node matching, and merge operations. This part has a time complexity of approximately $O(K \cdot (|E| \cdot d + N \cdot \log N))$, where $K$ is the number of scales, $|E|$ is the number of edges, $N$ is the number of nodes, and $d$ is the feature dimension. Secondly, in the network's forward propagation, the GCN encoder has a time complexity of $O((1+K) \cdot L \cdot |E| \cdot d_{\mathrm{hid}})$, where $L$ is the number of layers and $d_{\mathrm{hid}}$ is the hidden layer dimension. Thirdly, in the contrastive learning module, computing the similarities between node embeddings and the contrastive loss has a time complexity of $O(N^2 \cdot d_{\mathrm{hid}})$. Since it requires computing similarities between all pairs of nodes, this becomes the main bottleneck in large-scale graphs. Finally, in the self-supervised module, calculating the graph reconstruction loss and KL divergence loss has a time complexity of $O(N^2 \cdot d_{\mathrm{hid}} + N \cdot K \cdot d_{\mathrm{hid}})$. Combining the above analyses, the overall time complexity of the model can be expressed as: $O(K \cdot (|E| \cdot d + N \cdot \log N) + (1+K) \cdot L \cdot |E| \cdot d_{\mathrm{hid}} + N^2 \cdot d_{\mathrm{hid}})$. When the numbers of nodes and edges are large, the computations in the contrastive learning and self-supervised modules become the primary complexity bottlenecks.

Algorithm 1 shows the procedure of the MPCCL.

**Algorithm 1:** MPCCL for attributed graph clustering

**Input**: Graph data $G = (V, E, X; A)$, parameter $\lambda$, maximum iterations $N$, number of clusters $K$, scales of coarsened graphs $\{S_1,\ S_2,\ S_3\}$;

**Output**: Cluster partition results $\boldsymbol{r}$;

Initialize the parameters of GCN with pre-training;
Initialize **k** cluster centroids based on the representation from pre-trained auto-encoder;
Construct the multi-scale coarsened graphs from $G$ with scales $\{S_1,\ S_2,\ S_3\}$;
Create the optimizer with weight decay for L2 regularization;
**for** iter $\in$ 0, 1, $\boldsymbol{N}$ **do**
    Generate $x_{\text{aug}}$ by applying dropout on $X$ with a specified rate by Eq. (x);
    Extract coarsened graphs at each scale $G^{(s)} = \left\{G^{(s_1)}, G^{(s_2)}, \dots, G^{(s_K)}\right\}$ by Eq. (1);
    Obtain the representations $H^{(1)}$ by Eq. (14);
    Obtain multi-scale representations $H^{(2)}$ by Eq. (15);
    Compute the soft assignments $Q_1$ and $Q_2$ and $P$ by Eq. (29), Eq. (30) and Eq. (31);
    Feed the $H^{(1)}, H^{(2)}$ the decoder to reconstruct the raw data $X,\ A$;
    Calculate the objective function by Eq. (36);
    Optimize and update parameters of the whole model by back propagation;
**end**
Obtain the results of cluster partition $\boldsymbol{r}$ by Eq. (42);

# 4. Experiments

## 4.1 Benchmark Datasets

In our experiment, we utilized a total of five benchmark datasets. These datasets include ACM [41], DBLP [42], Cora [43], and Citeseer [44], all of which are graph-based datasets, while Reuters is a text-based dataset [45]. The detailed information for these five datasets is provided in Table 1.

**Table 1.** Details of the datasets

| Dataset | Type | Sample | Class | Dimension |
|---|---|---|---|---|
| ACM | Graph | 3025 | 3 | 1870 |
| DBLP | Graph | 4057 | 4 | 334 |
| Cora | Graph | 2708 | 7 | 1433 |
| Citeseer | Graph | 3327 | 6 | 3703 |
| Reuters | Text | 10000 | 4 | 2000 |

**ACM**: This dataset is divided into three categories: databases, data mining, and machine learning, containing 3,025 papers. Each paper is labeled with topic tags, and the content of the literature serves as the features of the papers.

**DBLP**: This dataset includes over 5.4 million academic papers classified into four categories: computer science, artificial intelligence, information retrieval, and data mining. It covers conference papers, journal articles, and preprints. Each paper has corresponding topic tags, and its field of study is regarded as its feature.

**Cora**: This dataset comprises 2,708 papers and 5,429 citation edges within seven different subject categories. Each paper is represented by a binary vector, indicating the specific topics the paper is associated with.

**Citeseer**: This dataset contains 3,327 papers divided into six distinct categories. It includes sparse feature vectors derived from the content of the papers, along with citation relationships among the papers.

**Reuters:** This text dataset consists of approximately 810,000 English news articles, which are primarily categorized into four main groups: Business and Industry, Government and Community, Markets, and Economy. We selected these four categories as labels. Due to computational constraints, we randomly chose a subset of 10,000 samples for the final clustering.

### 4.2 Evaluation criteria

To evaluate the clustering performance of the MPCCL model, we used four evaluation metrics: ACC (Accuracy), NMI (Normalized Mutual Information), ARI (Adjusted Rand Index), and F1 (F1 score), each assessing the model's classification effectiveness from different perspectives. ACC evaluates overall accuracy; NMI measures the alignment between clustering results and true labels; ARI assesses the consistency of clustering results with expected labels; and F1 balances precision and recall, reflecting the comprehensive performance of the classification. Higher values of these metrics indicate better clustering performance of the model.

### 4.3 Comparison methods

Table 2 summarizes the information on three typical types of graph clustering methods used in our comparative experiments, organized by “Method,” “Year,” “Advantages,” and “Disadvantages.” Each method was proposed in a specific year, with the “Advantages” column describing its strengths in handling particular tasks, while the “Disadvantages” column highlights its limitations.

**AE** [46]**.** The autoencoder retains the core features of the data through dimensionality reduction, enabling more efficient execution of clustering tasks with KMeans, thereby significantly enhancing clustering performance.

**GAE & VGAE** [7,47]**.** GAE learns low-dimensional node embeddings directly through GCN, avoiding the complexity of using probabilistic distributions and simplifying the variational inference process. Meanwhile, VGAE combines VAE with GCN, utilizing latent variables to learn the latent representation of undirected graphs. It employs GCN in the encoding phase and reconstructs the adjacency matrix through inner product in the decoding phase, allowing for efficient modelling and analysis of graph data.

**MBN** [48]**.** This attribute graph clustering model combines AE and GAE, improving clustering performance through feature complementarity and consistency constraints. The model optimizes representations and clustering assignments via a self-supervised module, enhancing accuracy and consistency in clustering.

**EGAE** [49]**.** Integrating GAE with relaxed k-means theory, it significantly improves clustering performance by optimizing the layout of graph representations in the inner product space.

**DFCN** [50]**.** This model integrates the representations generated by AE and an improved graph autoencoder (IGAE) through a designed Structure and Attribute Information Fusion (SAIF) module to learn a consistent representation. It adopts a target distribution generation method and utilizes a triple self-supervised learning strategy to leverage cross-modal information, thus enhancing clustering outcomes.

**DAEGC** [51]**.** Using a GAT encoder, it integrates both structural and content information of nodes into a compact embedding. Subsequently, a self-training mechanism optimizes clustering results, achieving mutual enhancement of embedding learning and clustering within a unified framework.

**ARGA** [52]**.** By incorporating adversarial regularization, the encoder-generated latent representations align with a preset prior distribution. This adversarial constraint effectively boosts the robustness and generalizability of graph embeddings.

**Table 2.** Summary of comparison methods

| Method | Year | Pros | Cons |
|---|---|---|---|
| AE<br>GAE & VGAE<br>DAEGC<br>ARGA<br>AGC<br>EGAE<br>DFCN<br>MBN | 1986<br>2016<br>2019<br>2019<br>2019<br>2021<br>2021<br>2023 | Enhances node representation richness and accuracy by combining node features with graph structural information. It exhibits strong scalability and adaptability, handling various types and scales of graph data, with excellent multi-task adaptability. | Primarily relies on the aggregation of local neighborhood information, making it challenging to capture dependencies between distant nodes effectively. This design limits the model's generalization and robustness when processing complex graph structures. |
| CONVERT<br>CLAGC<br>SCGC | 2023<br>2024<br>2025 | Enhances node representation quality through contrastive learning, achieving improved clustering performance with reduced dependency on labels, optimized computational efficiency, and excellent robustness. | Contrastive learning methods often use simple neighborhood selection to construct positive and negative sample pairs, overlooking dynamic changes within the graph structure. This static approach limits the model's ability to capture diverse features in complex graph structures. |
| TDCN<br>GC-SEE | 2022<br>2023 | Effectively enhances clustering accuracy and robustness through multi-scale feature fusion and self-supervised mechanisms. | Many existing methods focus excessively on low-order attribute information, neglecting high-order structural information, limiting the model's performance in capturing global graph features. |

**AGC** [53]**.** This model introduces an adaptive higher-order graph convolution method, capturing global cluster structures through high-order convolutions and adapting to diverse graph characteristics. It more effectively integrates structural information with node features.

**CONVERT** [54]**.** A graph contrastive clustering algorithm, it constructs a reversible perturbation-recovery network to generate enhanced views with reliable semantics. A label-matching mechanism is used to effectively mitigate semantic drift in enhanced views.

**CLAGC** [55]**.** A contrastive clustering algorithm for graph data, this model creates reliable contrastive views through a generative augmentation strategy and captures information from different feature spaces using a multi-level contrastive mechanism.

**TDCN** [56]**.** The proposed multi-scale Transformer dynamic fusion clustering network achieves effective aggregation and extraction of features across different data levels through multi-scale design.

**GC-SEE** [57]**.** By introducing a multi-scale feature fusion module and a self-supervised learning module, it fully exploits node importance, attribute importance, and multi-level structural information to improve clustering performance.

**SCGC** [58]**.** By introducing graph structures into the contrast loss signal, discriminative node representations can be learned and soft clustering labels can be iteratively optimized to improve the clustering effect by directly exploiting the graph structure information. This app roach eliminates the need for adjacency matrices or message passing and enables node embedding by virtue of contrast learning constraints alone, while significantly reducing the consumption of computational resources.

### 4.4 Implementation details

**Table 3.** Configuration of model parameters for the five benchmark datasets

| Dataset | Mask | Regularization Coefficient | Scale 1 | Scale 2 | Scale 3 | Weight Decay | Learning Rate |
|---|---|---|---|---|---|---|---|
| ACM | 0.4 | 0.1 | 0.5 | 0.25 | 0.1 | 1e-3 | 0.0005 |
| DBLP | 0.2 | 0.05 | 0.3 | 0.15 | 0.06 | 1e-6 | 0.0005 |
| Citeseer | 0.3 | 0.1 | 0.2 | 0.1 | - | 1e-2 | 0.0005 |
| Cora | 0.1 | 0.0005 | 0.2 | 0.1 | - | 1e-2 | 0.0005 |
| Reuters | 0.2 | 0.1 | 0.2 | 0.05 | - | 1e-2 | 0.0005 |

In the preparation phase for model training, we conducted 50 rounds of pre-training on the GCN component to obtain high-quality initial feature representations, retaining the training weights to enhance the initial clustering effectiveness. During pre-training, each encoding layer of the GCN was set to dimensions of 256 and 512, with symmetric settings for each decoding layer. The MLP layer's dimension was configured as 1024. Specifically, we fixed the learning rate at 0.005. The mask is tuned from 0.1 to 0.8; the regularization coefficient is searched in (0.0005,0.001,0.005,0.01,0.05,0.1), the weight decay is tuned from 1e-6 to 1e-2 and scale settings for each dataset are shown in Table 3. After pre-training, we applied the KMeans clustering algorithm to the final feature representations, repeating the clustering 20 times to select the optimal result. In the main training phase, the model performs multi-scale feature learning on both the original and coarsened graphs, with the multi-scale settings tailored to each dataset as indicated in table 3. The total number of training epochs was fixed at 200. All experiments were conducted with a random seed to ensure reproducibility. The model was run in an environment equipped with an Intel i5-1240P CPU, NVIDIA GTX 3070 GPU, and 32 GB of RAM, and deployed on a Windows 10 system based on PyTorch 3.9.0.

### 4.5 Clustering Results

In our clustering experiments as shown in table 4, our approach demonstrates significant advantages over three typical types of graph clustering algorithms across five datasets, especially in achieving substantial performance improvements on multiple evaluation metrics. Firstly, the first type of clustering methods based on jointly leveraging graph structural information and node features (e.g., AE, GAE & VGAE, MBN, EGAE, DFCN, DAEGC, ARGA, and AGC) enhance node embeddings by aggregating local neighborhood information. These methods often show limitations in generalization when handling complex graph structures due to their insufficient capability in capturing long-range node relationships. For example, on the Cora dataset, the ACC of GAE and VGAE is 63.80% and 64.34%, respectively, while our model, by utilizing multi-scale graph coarsening, effectively captures global information and integrates long-range associations between nodes. On the ACM dataset, our method improves the NMI metric by 5.55% compared to the next best method, MBN. Our model achieves a notable improvement in clustering performance through global regularization and multi-scale information fusion. For instance, on the Cora dataset, our model outperforms DAEGC and DFCN in ACC by 6.36% and 44.26%, respectively.

Secondly, the second type of contrastive learning-based clustering methods (e.g., CONVERT and CLAGC) enhance the stability and robustness of clustering results by optimizing node embeddings through contrastive learning mechanisms. However, these methods often use static neighborhood relationships to select positive and negative sample pairs, overlooking dynamic changes in graph structures. For example, CLAGC achieves an ARI of 75.97% on the ACM dataset but constructs sample pairs in a static manner, failing to fully capture the diversity of complex graph structures. In contrast, our method, using a dynamic multi-scale selection mechanism for positive and negative samples, leverages multi-layered information in constructing sample pairs and adopts cluster centroids as positive samples, further enhancing the model's discriminative capability. On the ACM dataset, our model improves the F1 metric by 3.68% compared to CLAGC.

Lastly, the third type of multi-scale graph clustering methods (e.g., TDCN and GC-SEE) capture information at different levels through multi-scale feature fusion, enhancing adaptability to multi-level graph information. However, these methods often focus on low-order structural information and neglect the integration of global information. For example, GC-SEE achieves an NMI of 73.99% and an ACC of only 69.79% on the Cora dataset. On the same dataset, our approach combines multi-scale graph coarsening with contrastive learning, enabling the model to integrate high-order information from multi-layer perspectives while maintaining effective global structure capture. This multi-scale design shows remarkable performance on complex datasets, with our method improving the ACC on the Cora dataset by 4.25% compared to GC-SEE. Our approach shows significant advantages in capturing long-range dependencies within graphs, dynamic positive and negative sample selection, and global information integration. Through the innovative design of multi-scale regularization and contrastive learning, our approach demonstrates outstanding clustering performance across five benchmark datasets, further advancing the frontiers of research in graph clustering.

**Table 4.** The clustering results of 13 comparison methods and our method across five datasets.

| Dataset | Metric | AE | GAE | VGAE | EGAE | DAEGC | ARGA | AGC | DFCN | MBN | CONVERT | CLAGC | TDCN | SCGC | OURS |
|---|---|---|---|---|---|---|---|---|---|---|---|---|---|---|---|
| ACM | ACC | 82.35±0.08 | 84.51±1.42 | 84.11±0.21 | 83.63±3.97 | 85.28±0.27 | 64.65±1.93 | 79.64±0.61 | 90.70±0.31 | 93.01±0.13 | 86.35±1.54 | 91.31±0.24 | 90.82±0.19 | 92.20±0.55 | **94.68±0.39** |
| | NMI | 48.73±0.16 | 55.39±1.91 | 53.20±0.51 | 52.00±7.92 | 55.01±0.83 | 30.75±2.10 | 48.82±0.93 | 69.27±0.66 | 74.94±0.26 | 58.35±3.50 | 69.40±0.67 | 69.20±0.50 | 72.89±0.31 | **79.98±1.13** |
| | ARI | 54.85±0.16 | 59.48±3.10 | 57.71±0.69 | 57.83±8.71 | 61.20±0.64 | 20.20±3.47 | 50.59±1.05 | 74.62±0.75 | 80.39±0.33 | 63.89±3.71 | 75.97±0.50 | 74.84±0.46 | 78.31±0.72 | **84.85±1.03** |
| | F1 | 82.19±0.08 | 84.68±1.31 | 84.19±0.22 | 34.22±26.6 | 85.31±0.28 | 64.01±1.99 | 80.08±0.59 | 90.68±0.31 | 93.02±0.13 | 86.33±1.52 | 91.33±0.22 | 90.80±0.19 | 92.13±0.47 | **94.69±0.40** |
| Cora | ACC | 49.93±0.17 | 63.80±1.29 | 64.34±1.34 | 68.97±3.68 | 67.72±0.19 | 59.44±4.50 | 64.90±2.35 | 49.93±0.08 | 69.79±0.57 | 73.99±1.54 | **75.08±0.04** | - | 47.64±1.43 | 72.03±0.52 |
| | NMI | 26.59±0.23 | 47.64±0.37 | 48.57±1.09 | 50.27±3.06 | 49.98±0.11 | 41.02±2.22 | 51.67±1.56 | 26.59±0.16 | 48.81±1.03 | 55.50±1.11 | **59.21±0.10** | - | 35.08±1.31 | 53.86±0.48 |
| | ARI | 21.94±0.16 | 38.00±1.19 | 40.35±1.46 | 44.61±4.04 | 43.39±0.19 | 34.27±3.29 | 41.05±3.42 | 21.94±0.16 | 47.03±0.88 | 50.49±2.03 | **56.59±0.07** | - | 19.14±0.91 | 52.29±0.80 |
| | F1 | 48.38±0.24 | 65.86±0.69 | 64.03±1.34 | 15.73±9.79 | 66.12±0.18 | 59.57±4.16 | 68.63±4.14 | 48.38±0.08 | 64.89±0.63 | 72.84±3.26 | **73.00±0.05** | - | 27.44±0.78 | 69.73±0.44 |
| CiteSeer | ACC | 62.37±0.13 | 61.38±0.80 | 60.98±0.38 | 52.83±5.12 | 66.55±0.11 | 42.65±3.27 | 67.87±0.10 | 69.37±0.15 | 68.19±0.34 | 67.85±0.65 | 69.43±0.17 | 69.10±0.53 | 69.55±0.26 | **70.56±0.79** |
| | NMI | 35.50±0.08 | 34.62±0.68 | 32.70±0.29 | 28.33±3.40 | 41.61±0.09 | 20.08±2.79 | 42.24±0.10 | 43.21±0.19 | 42.56±0.67 | 41.02±0.85 | 44.31±0.20 | 41.61±0.65 | 44.84±0.29 | **45.10±0.63** |
| | ARI | 34.99±0.14 | 33.58±1.19 | 33.12±0.52 | 25.62±5.05 | 42.58±0.09 | 14.45±2.38 | 43.03±0.17 | 45.12±0.18 | 45.02±0.52 | 42.01±0.12 | 45.82±0.31 | 43.91±0.56 | 45.79±0.72 | **46.90±1.12** |
| | F1 | 60.02±0.11 | 57.38±0.81 | 57.70±0.49 | 16.86±8.02 | 62.59±0.15 | 41.80±3.48 | 63.86±0.08 | 64.69±0.17 | 63.67±0.27 | 62.17±1.88 | 65.36±0.16 | 62.30±0.31 | 65.87±0.21 | **66.09±0.66** |
| DBLP | ACC | 66.85±0.35 | 61.20±1.21 | 58.59±0.08 | 53.83±9.17 | 65.96±0.18 | 46.09±3.84 | 64.27±0.31 | 76.24±0.21 | 77.04±0.06 | 63.75±2.29 | 81.31±0.38 | - | 77.30±0.08 | **81.89±0.20** |
| | NMI | 34.49±0.16 | 30.80±0.91 | 26.91±0.08 | 21.57±7.44 | 32.03±0.25 | 12.44±2.09 | 33.59±0.35 | 44.02±0.24 | 46.95±0.11 | 27.92±2.51 | 52.36±0.71 | - | 45.33±0.25 | **52.58±0.40** |
| | ARI | 29.19±0.91 | 22.01±1.40 | 17.91±0.09 | 17.54±10.4 | 34.76±0.23 | 10.30±2.46 | 28.83±0.27 | 46.95±0.50 | 51.01±0.11 | 27.07±2.99 | 57.77±0.87 | - | 49.85±0.33 | **58.37±0.39** |
| | F1 | 67.54±0.06 | 61.41±2.21 | 58.70±0.09 | 22.70±10.1 | 64.19±0.22 | 45.99±3.90 | 68.27±0.33 | 76.04±0.19 | 76.34±0.06 | 63.78±2.32 | 80.76±0.36 | - | 76.63±0.07 | **81.47±0.22** |
| Reuters | ACC | 74.90±0.21 | 54.40±0.27 | 60.85±0.23 | 59.32±6.31 | 60.27±0.30 | 49.66±5.08 | 50.35±7.36 | 77.76±0.31 | 77.89±0.17 | 70.85±1.33 | 64.33±5.39 | 81.50±1.09 | 73.49±2.39 | **82.48±1.40** |
| | NMI | 49.69±0.29 | 25.92±0.41 | 25.51±0.22 | 31.90±8.54 | 33.04±0.09 | 20.64±5.30 | 30.04±8.48 | 59.82±0.80 | 50.58±0.38 | 43.08±2.10 | 42.06±4.39 | 59.28±0.97 | 46.32±1.42 | **59.90±0.97** |
| | ARI | 49.55±0.37 | 19.61±0.22 | 26.18±0.36 | 30.68±9.20 | 30.31±0.59 | 18.24±5.06 | 17.19±9.09 | 59.76±0.80 | 56.83±0.39 | 43.41±2.65 | 35.48±7.75 | 62.46±1.94 | 52.83±0.43 | **63.12±2.00** |
| | F1 | 64.25±0.22 | 43.53±0.42 | 57.14±0.17 | 22.65±13.2 | 57.45±0.14 | 46.12±5.31 | 36.90±8.07 | 69.57±0.25 | **70.48±0.17** | 66.96±1.47 | 54.07±6.00 | 66.68±0.23 | 65.85±3.12 | 64.86±0.51 |

Clustering performance (%) of our method and 13 comparison methods.

**Bold** values are the best results, underlined values are the second-best results.

"-" indicates that the method is not mentioned.

### 4.6 Parameter Analysis

In this section, we conduct a sensitivity analysis of parameters on the ACM, Citeseer, DBLP, Cora and Reuters datasets.

**Parameters of different scales**

By visualizing the impact of parameter changes on model performance, we can intuitively observe how scale at large, medium, and small levels and the regularization coefficient influence clustering performance. The scale parameter range in this study is as follows: we consider two different scale ratio schemes. In the dual-scale scheme, ratios are set to $s$ and $s + 0.1$, where $s \in [0.1, 0.2, 0.3, 0.4]$; For the large dataset Reuters, after experimenting with the scale settings of $s$ and $s + 0.1$, we found that the performance of Reuters under these scale ratios was suboptimal. Therefore, we explored smaller scale ratios with $s$ and $r$, where $r \in [0.01, 0.05, 0.1]$; in the triple-scale scheme, ratios are set to $s$, $0.5s$, and $0.2s$, with $s \in [0.3, 0.5, \ldots, 1]$. The choice of the regularization coefficient is also crucial. According to Raphael, an overly large L2 norm can lead to underfitting, resulting in information loss [59], while an overly small L2 norm may lead to overfitting, manifesting as high accuracy during the training phase but significant fluctuation during the testing phase.

**Parameters of different mask**

As shown in Figure 5, the ACM dataset exhibits the most stable performance across different Mask parameters, with an average ACC consistently around the high-performance level of 0.94. The DBLP dataset achieves an average ACC of 0.8189 when the Mask is set to 0.2, after which the performance gradually declines. The Cora dataset has an average ACC of 0.7203, showing a downward trend as the Mask parameter increases. The Citeseer dataset’s performance remains relatively steady but lower, with an initial average ACC of approximately 0.7056, experiencing slight fluctuations as Mask increases. Reuters achieves an average ACC of 0.8248 when the mask is set to 0.2, with the ACC value showing minimal fluctuations when the mask is greater than 0.1 and less than 0.7. This indicates that moderate masking can enhance model generalization, while excessive masking may harm clustering performance.

**Parameters of different regularization coefficients**

We employ Normalized Mutual Information (NMI) as the evaluation metric to investigate the demand for regularization parameters across diverse datasets and their corresponding response characteristics. As illustrated in Figure 6, the influence of the regularization coefficient on NMI exhibits pronounced heterogeneity among the datasets. For the ACM dataset, NMI attains its peak value of 0.8107 when the regularization coefficient is elevated to 0.1, indicating that robust regularization constraints effectively optimize the separability of its latent feature space. Similarly, the Citeseer dataset achieves optimal NMI at a regularization coefficient of 0.1, further corroborating the positive regulatory effect of high regularization intensity on the model's generalization capacity. Notably, the Cora, Reuters, and DBLP datasets demonstrate remarkable robustness to adjustments in the regularization coefficient. Specifically, Cora's NMI values remain highly stable across a broad range of regularization coefficients, from 0.0005 to 0.1, with a negligible standard deviation of merely 0.0002. Reuters maintains its NMI consistently around 0.8, while DBLP confines its NMI fluctuations within a narrow range of 0.7876 to 0.7906, even as the regularization coefficient escalates from 0.0005 to 0.1. This phenomenon suggests that our model, without relying on meticulous regularization tuning, can still capture the intrinsic structural features of the datasets through an adaptive mechanism, thereby achieving a dynamic equilibrium between complexity and generalizability.

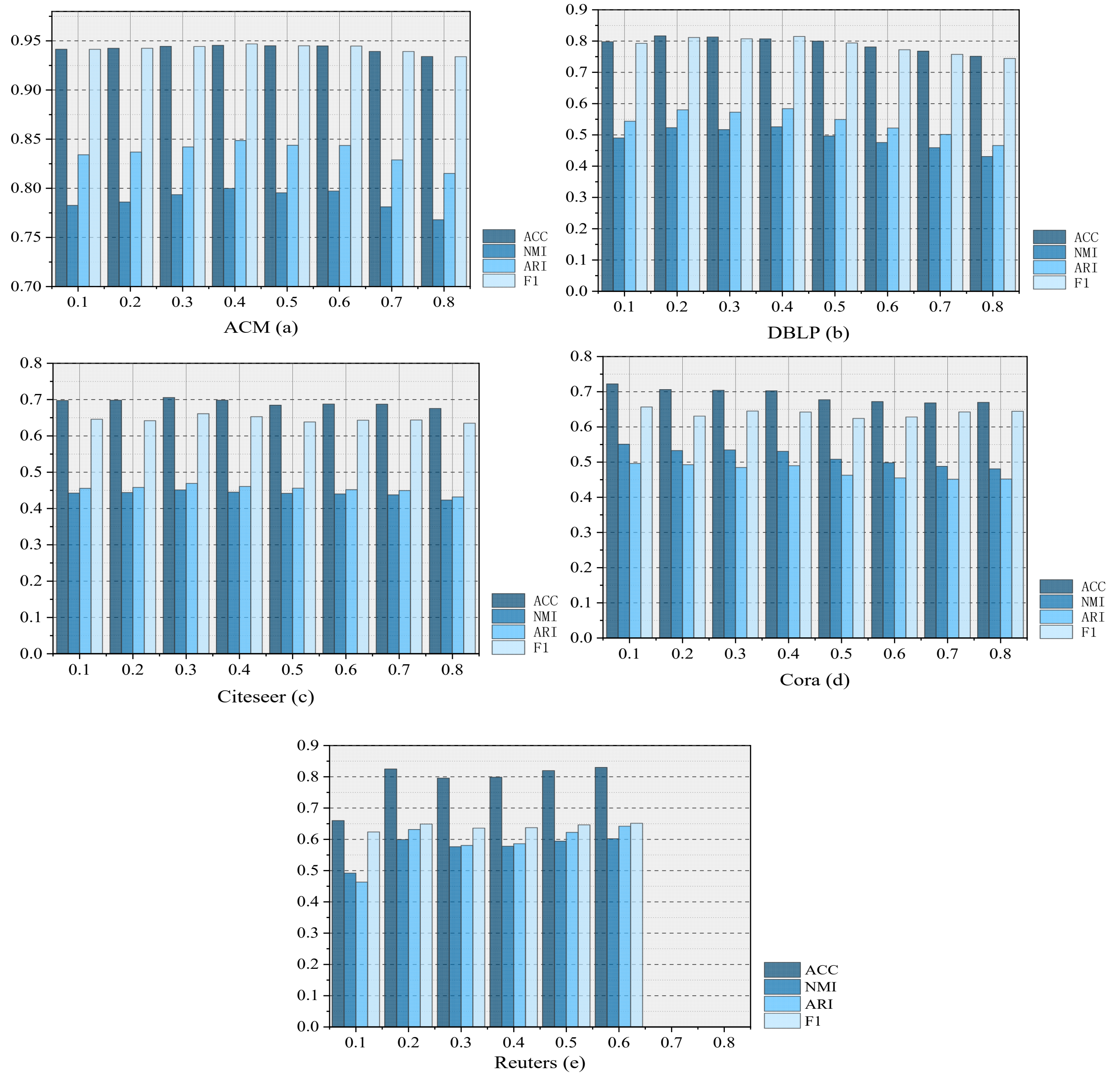

*The 0.7 and 0.8 scales of the Reuters dataset cannot be clustered

**Fig. 5.** The model's performance across five datasets under different Mask parameters. In this case, Reuters fails to cluster when mask ∈ [0.7, 0.8].

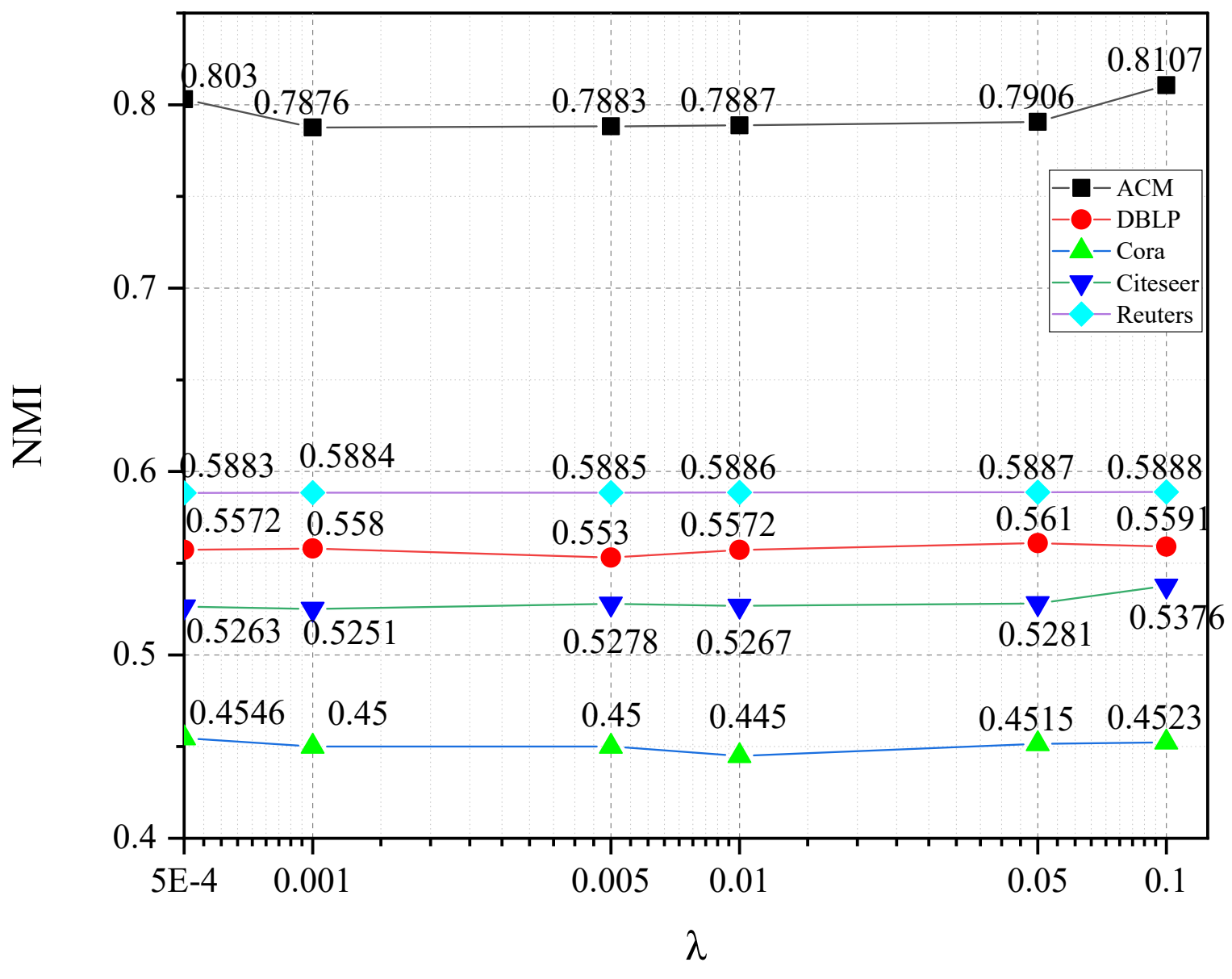

**Fig. 6.** Model performance across five datasets under different Regularization Coefficient parameters

**Different weight decay parameters**

Regarding the L2 regularization parameter weight decay, this hyperparameter effectively constrains the rapid growth of model parameters by adding a squared parameter penalty term to the loss function during optimization, thereby reducing model complexity and mitigating overfitting risks. As evidenced by the experimental results in Figure 7, the optimal weight decay for the ACM dataset is 1e-3, demonstrating that moderate regularization not only suppresses overfitting but also preserves the model's capacity to learn complex patterns.

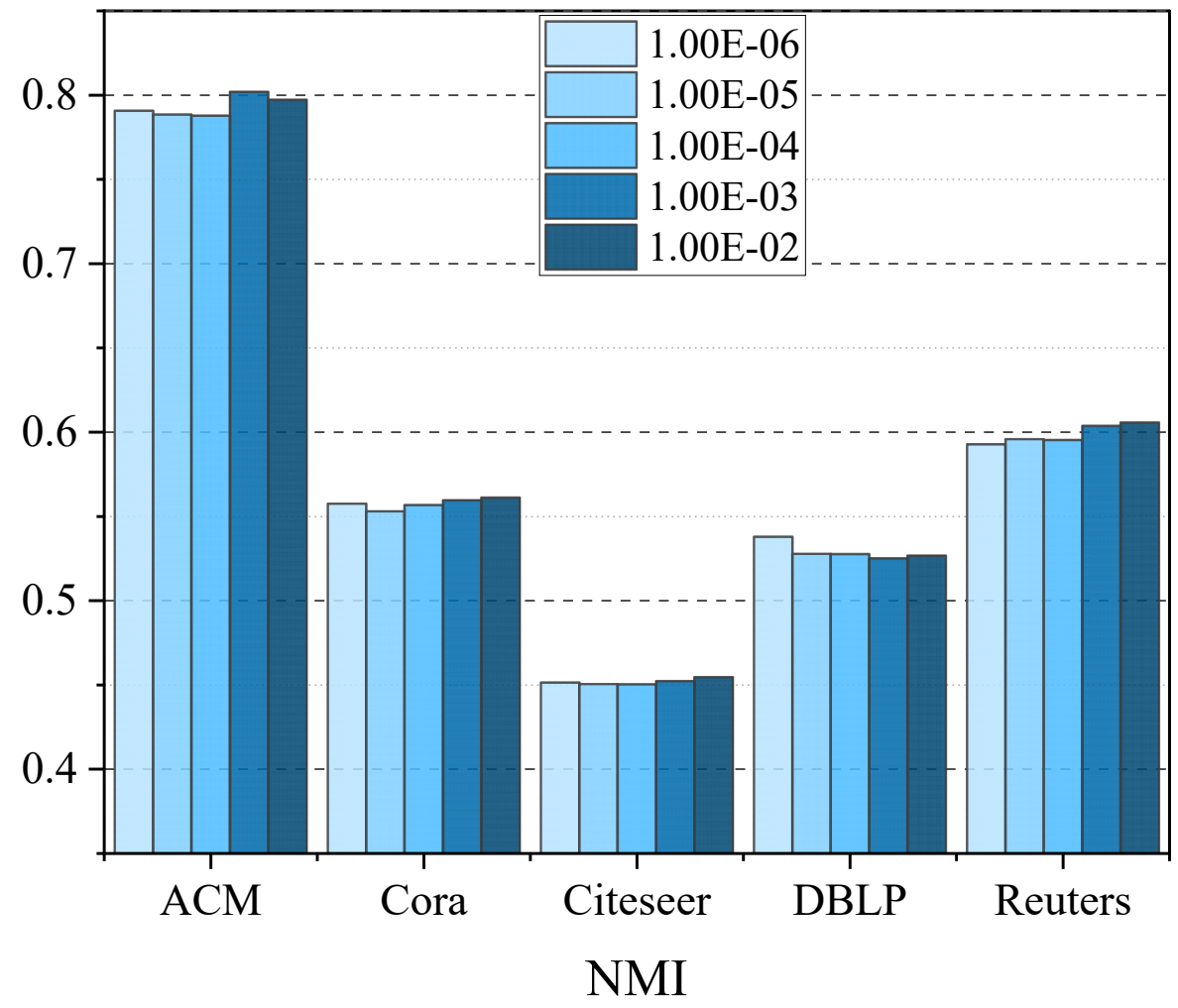

**Fig. 7.** Model performance across five datasets under different weight decay parameters

The DBLP model achieves its best performance at a lower regularization strength of 1e-6, indicating that excessively strong constraints may hinder the full extraction of latent semantic features. In contrast, the Citeseer and Reuters datasets attain peak performance at a weight decay of 1e-2, suggesting that stronger regularization helps suppress redundant feature interactions while maintaining the model's generalization boundaries. Although the Cora dataset reaches its optimal performance at the same weight decay 1e-2 as Citeseer and Reuters, this is attributed to Cora's high feature dimensionality and intricate local structural complexity, which predispose the model to overfitting. The more aggressive regularization effectively curbs redundant interactions among high-dimensional features while preserving critical ones, thereby enhancing the model's generalization capability.

Table 5 indicates that the model performs best at smaller scales, particularly in the Cora and Reuters dataset, suggesting that the multi-scale approach effectively captures local structures and reduces noise. In contrast, the ACM, DBLP, and Citeseer datasets achieve optimal performance at a medium scale, likely due to the complexity of their data, where a balanced scale better represents their features. As the scale increases, the effectiveness of the multi-scale approach diminishes, resulting in a decline in model performance. Additionally, existing research commonly employs a fixed learning rate. To ensure stable convergence and mitigate excessive fluctuations in parameter updates, we adopt a constant learning rate across all five datasets.

As shown in Figure 8, these datasets exhibit densely packed, tightly connected nodes in the central areas of the original graph, reflecting strong intra-cluster associations within core topics, while the peripheral nodes are sparser, representing marginal or secondary topics. After multi-scale coarsening, the graph structure of each dataset is simplified, with peripheral nodes and secondary connections gradually reduced. This clarification of key relationships among primary topic clusters and core themes highlights the structure and associations of the core topics within each dataset. This process effectively removes non-core nodes and secondary connections, making the analysis of thematic relationships more intuitive and easier to understand.

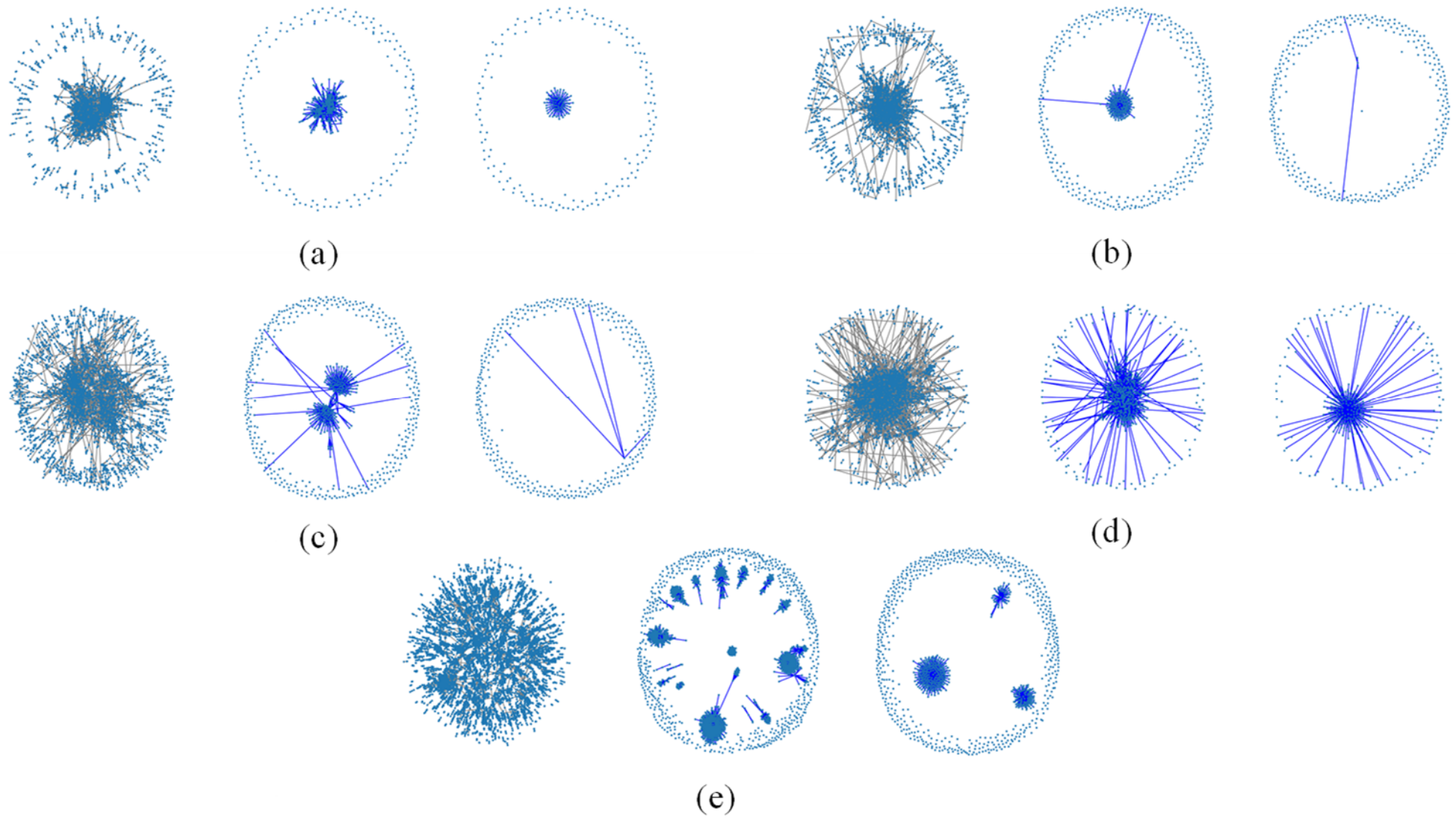

**Fig. 8.** Structural visualization under different coarsening scales for the five datasets. In the Figure, (a) shows the ACM dataset with coarsening scales [1, 0.5, 0.25, 0.1], (b) shows the DBLP dataset with coarsening scales [1, 0.3,

0.15, 0.06], (c) shows the Citeseer dataset with coarsening scales [1, 0.2, 0.1], and (d) shows the Cora dataset with coarsening scales [1, 0.2, 0.1]. (e) shows the visualization of the graph for the Reuters dataset at the coarsening scales of [1, 0.2, 0.05].

**Table 5.** The optimal scale and regularization coefficient

| Dataset | Scale 1 | Scale 2 | Scale 3 | Regularization Coefficient |
|---|---|---|---|---|
| **ACM** | 0.5 | 0.25 | 0.1 | 0.1 |
| **Citeseer** | 0.1 | 0.2 | - | 0.1 |
| **DBLP** | 0.3 | 0.15 | 0.06 | 0.1 |
| **Cora** | 0.1 | 0.2 | - | 0.1 |
| **Reuters** | 0.2 | 0.05 | - | 0.1 |

### 4.7 Ablation Analysis

**Experiment 1. Comparison between Single Scale and Multi-Scale**

As shown in Figure 9, using multiple scales in our model generally outperforms single-scale usage across all datasets. This is because the multi-scale model, by integrating information from various scales, provides a more comprehensive feature representation, thereby enhancing the model's generalization ability. In contrast, the single-scale model, lacking sufficient information, struggles to achieve the same level of effectiveness.

**Experiment 2. Ablation Study on One-to-Many Contrastive Learning Module**

The heatmaps in Figure 10 illustrate the similarity distributions for different positive sample selection strategies in contrastive learning on the ACM dataset: (a) The similarity matrix using the one-to-many positive sample selection strategy reveals a distinct block structure, reflecting several high-similarity groups, which is suitable for capturing broad group similarity; (b) The similarity matrix with the one-to-one positive sample selection strategy shows a more dispersed similarity distribution while retaining group characteristics, indicating that this strategy captures finer-grained similarity information; (c) The difference matrix displays the similarity differences between the two strategies, with color variations indicating specific node pairs where the similarity assessments differ. The one-to-many strategy excels in capturing intra-class similarity and group aggregation compared to the one-to-one strategy. By selecting multiple positive samples, the one-to-many strategy effectively enhances the similarity among intra-class nodes, resulting in tighter clustering, making it particularly suitable for clustering tasks that require stronger intra-class consistency. Although the one-to-one strategy has advantages in reducing redundant information, it is relatively weaker in capturing complex intra-class structures.

In our model, one-to-many contrastive learning consistently outperforms one-to-one contrastive learning as shown in table 6. This is primarily because one-to-many contrastive learning leverages multiple positive samples for comparison, providing richer information and context, thereby enhancing the model's understanding of inter-sample similarity and improving its robustness and generalization capability.

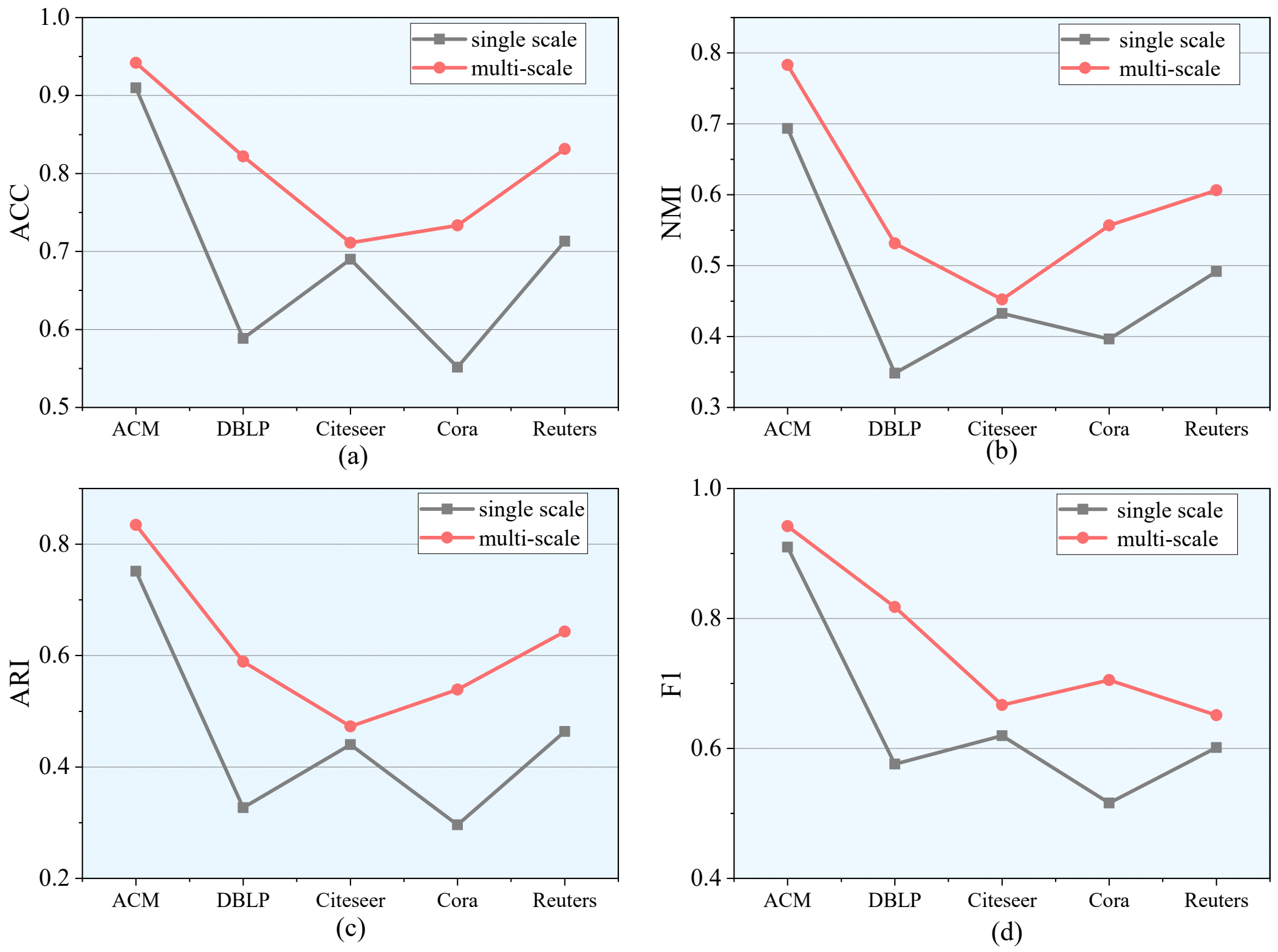

**Fig. 9.** Comparison of Single Scale and Multi-Scale Models across Five Datasets

**Table 6.** Performance of One-to-Many and One-to-One Contrastive Learning on ACC Metrics across Five Datasets

| **Dataset** | **ACM** | **DBLP** | **Cora** | **Citeseer** | Reuters |
|---|---|---|---|---|---|
| One-to-Many | **94.68** | **81.89** | **72.03** | **70.56** | **83.15** |
| One-to-One | 92.51 | 80.07 | 66.69 | 65.23 | 79.13 |

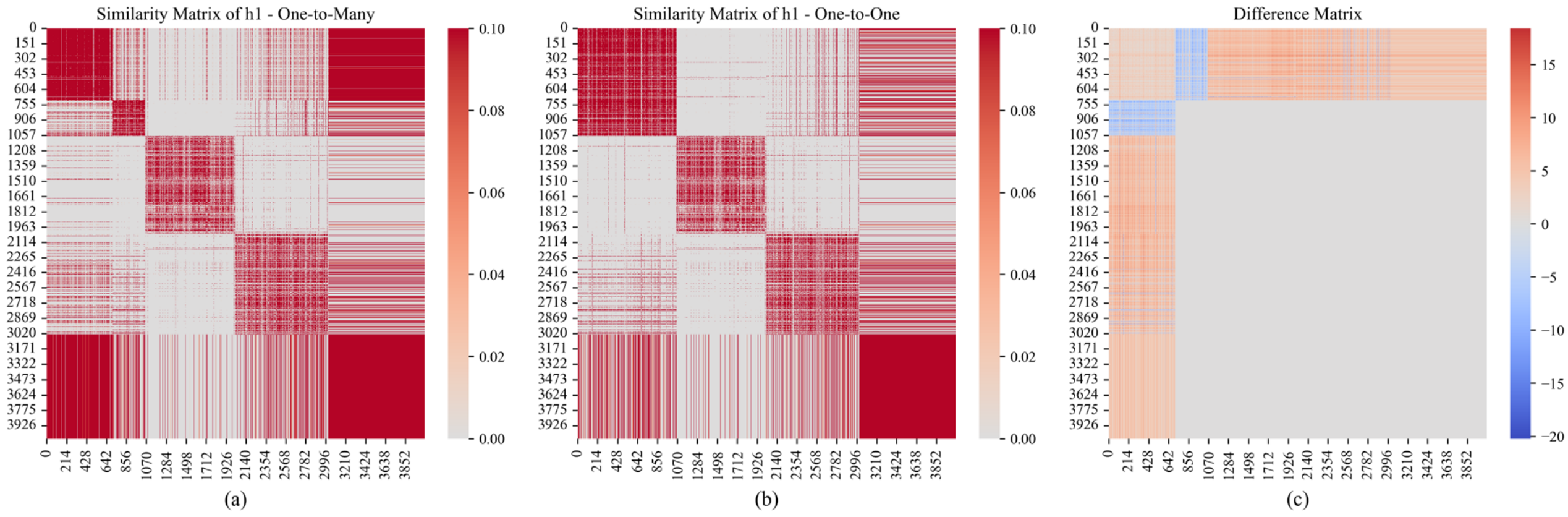

**Fig. 10.** Embedding Representations Produced by One-to-Many and One-to-One Models on the ACM Dataset

**Experiment 3. Comparison of Model Performance between Single-View and Dual-View**

As shown in Figure 11, using dual-view contrastive learning in the contrastive learning module consistently outperforms the single-view approach across all datasets. The dual-view model achieves both higher average and peak accuracy across the five datasets compared to the single-view model. We attribute this to the dual-view model's ability to leverage the rich information and diverse positive and negative sample pairs provided by different perspectives, enabling the model to better capture complex feature relationships and enhance generalization capability. In contrast, the single-view model is limited by its single-source information and lack of sample diversity, making it challenging to achieve comparable results.

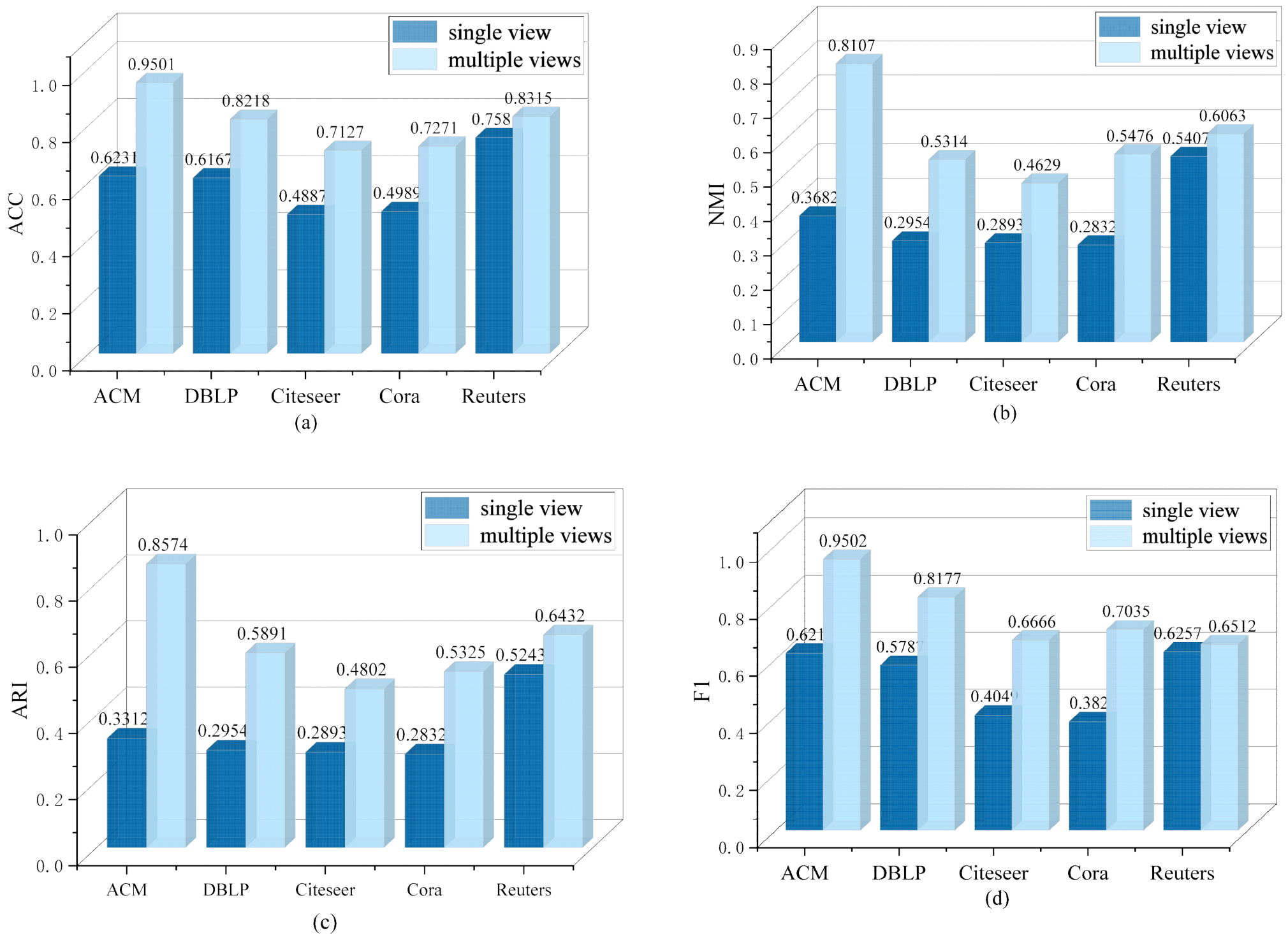

**Fig. 11.** Performance of Single-View and Dual-View Models across Five Datasets

**Experiment 4. Comparison between Models with and without Regularization**

Figure 12 shows that the model with a regularization term achieves a higher average performance across all datasets compared to the model without regularization. We believe that by computing the normalized Laplacian matrix, the regularization term leverages the structural information of the graph, enhancing the modeling of relationships between nodes and encouraging the formation of more compact clusters in the embedding space, thereby improving stability.

## 4.8 Time complexity analysis

To ensure the accuracy and comparability of the experimental results, we run them in a unified experimental environment to ensure that the single variable is the algorithm itself. We chose as a test benchmark the most challenging dataset, Reuters, which has a complexity that fully reflects the difference in performance of different methods when dealing with large-scale data. By comparing the running times of the methods on this dataset in Figure 13, we are able to visually assess the advantages of the proposed methods in terms of computational efficiency.

## 4.9 GPU memory analysis

Also, in our experiments, we have analyzed the GPU memory usage of the proposed method in comparison with the other four methods on the ACM dataset. Table 7 shows that when the ACM dataset is selected, our method performs at a disadvantage compared to the other four methods in terms of GPU memory footprint. We realize that this is a key issue that needs to be improved. In our future research, we will work on optimizing the memory management mechanism of the algorithm and exploring more efficient memory utilization strategies to reduce the GPU memory footprint, so as to improve the applicability and competitiveness of our method on different datasets.

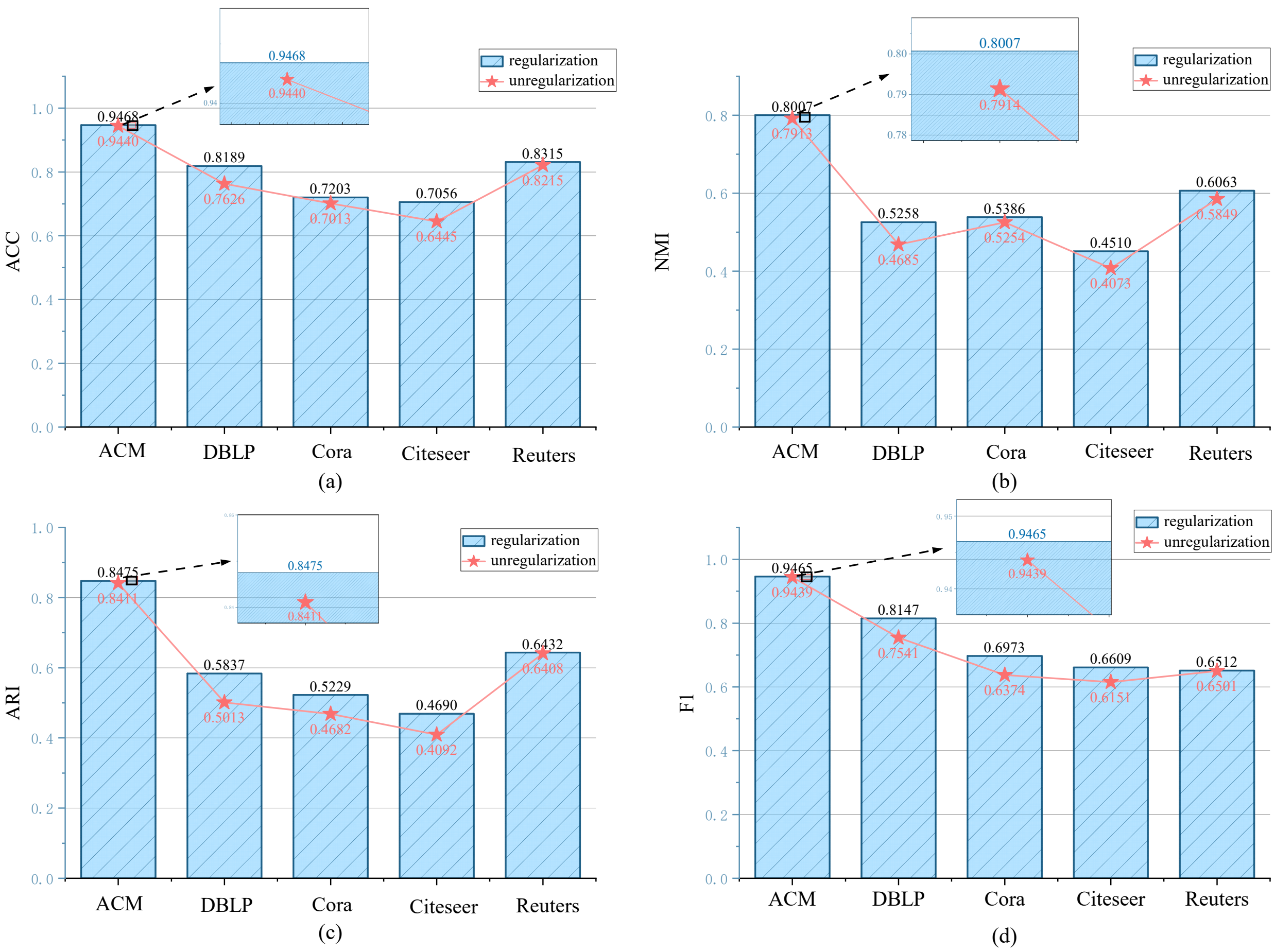

**Fig. 12.** Performance of Models with and without Regularization across Four Metrics on Five Datasets

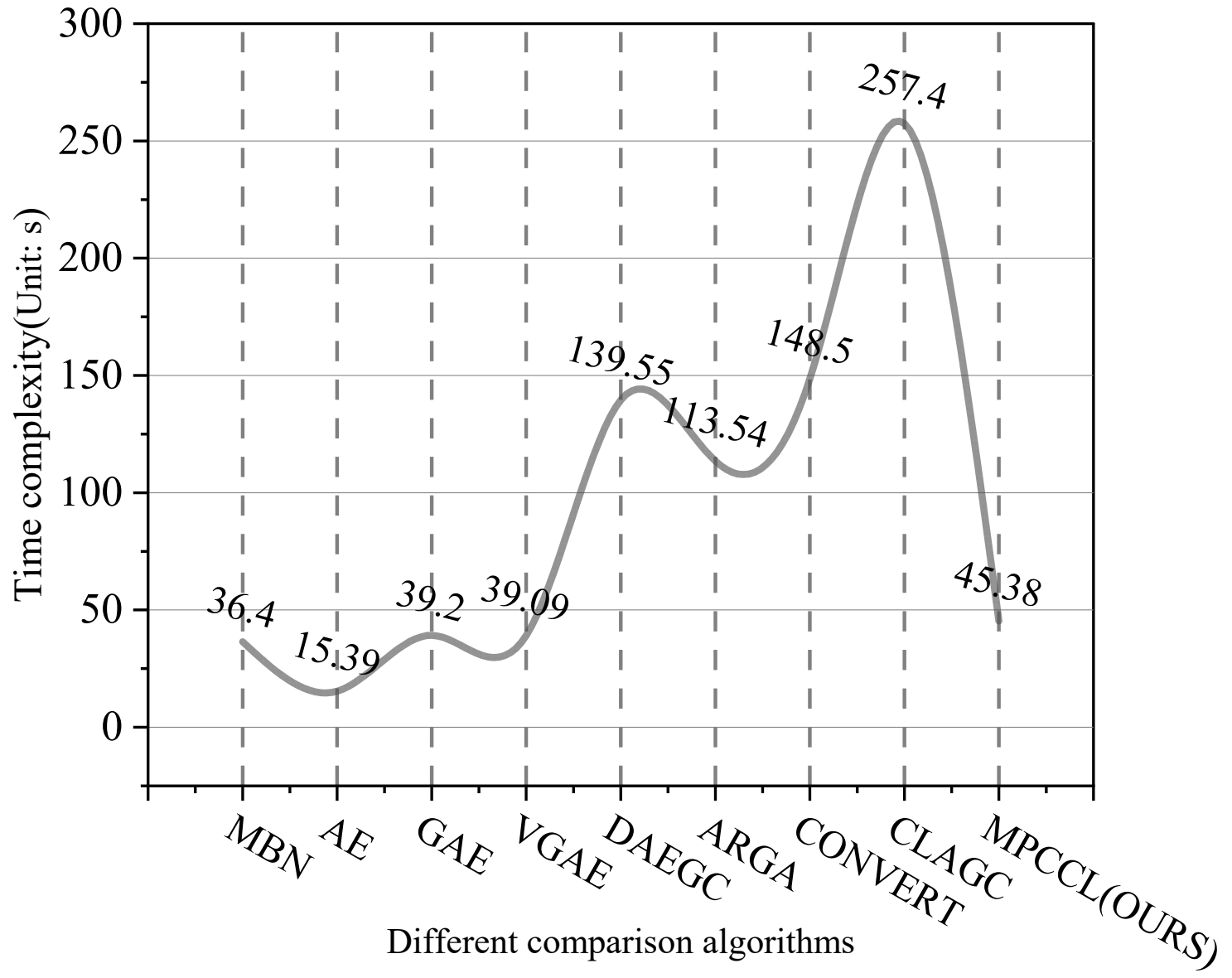

**Fig. 13.** The time complexity of our model compared with the other eight models

**Table 7.** GPU memory usage comparison

| Methods | AE | MBN | SCGC | CLAGC | Ours |
|---|---|---|---|---|---|
| GPU memory | 466 MiB | 914 MiB | 781 MiB | 811 MiB | 1789 MiB |

### 4.10 Statistical Significance Test

To ascertain the statistical significance of our method's performance gains, we employed a standard threshold of $\log_{10}(0.05) \approx -1.30$. For the ACM and Reuters datasets—where the publicly released MBN code represents the strongest available baseline—we performed two-sided paired t-tests between our method and MBN for each metric; on Citeseer and DBLP, we used the same test against CLAGC. As shown in Figure 14, every experimental scenario produced p-values well below $\log_{10}(0.05) \approx -1.30$, compellingly demonstrating that our observed improvements cannot be ascribed to random variation but instead reflect highly reproducible, systematic advances.

Although the absolute gains in NMI and ARI may appear modest, our method's synergistic fusion of contrastive learning with graph coarsening introduces a "one-to-many" contrastive mechanism. By treating cluster centroids as positive anchors, we align features of both high- and low-frequency nodes within each cluster—mitigating bias toward high-frequency nodes while amplifying learning signals for underrepresented ones. This enhancement of feature discriminability translates into a statistically robust stability advantage in clustering performance. Consequently, the results substantiate our method's superiority over both MBN (on ACM and Reuters) and CLAGC (on the other datasets) from dual perspectives: algorithmic innovation and rigorous statistical validation.

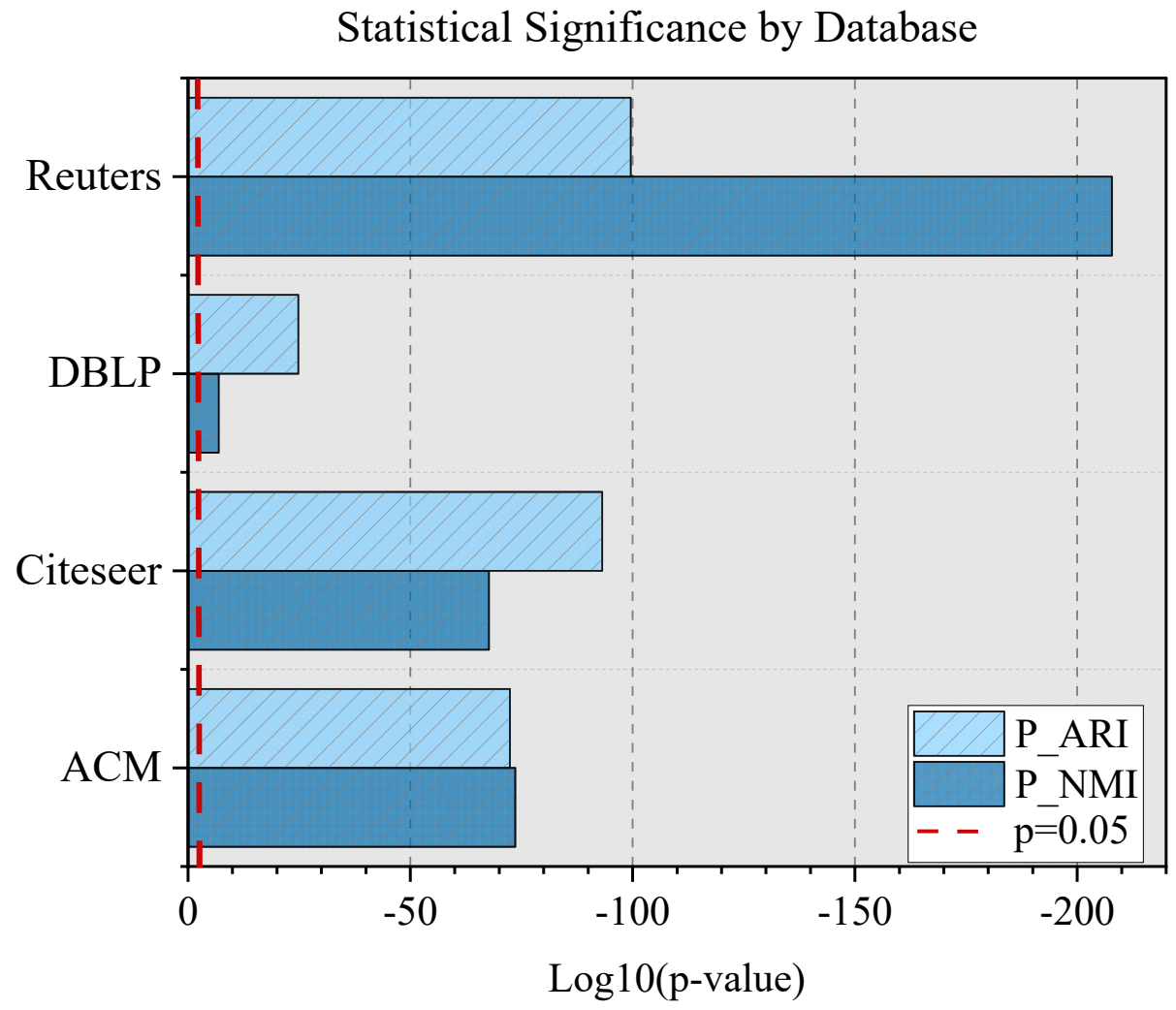

**Fig. 14.** Statistical Significance of Clustering Performance—MBN on ACM & Reuters; CLAGC on the Other Two Datasets

## 5. Discussion

Cora dataset contains a large number of nodes and complex reference relationships, and presents a typical long-tail distribution feature, that is, the degree of nodes is uneven, and most nodes are only connected to a small number of neighbors, while a few nodes are highly connected. This highly heterogeneous structural property leads to long and unstable information propagation paths between nodes (reducing information propagation efficiency), which restricts the effective capture of long-distance node dependencies by graph convolutional networks (such as GCN) based on local neighborhood information aggregation. Specifically, when aggregating neighborhood information, standard GCN treats all neighbors equally by default, and does not distinguish between core and edge nodes, which is particularly bad in long-tail graph structures (such as Cora data sets) and may result in inadequate generalization and over smoothing of node representations [60]. In addition, although the multi-scale graph coarsening technique we adopted effectively reduces the complexity of the graph by merging nodes layer by layer, it also has obvious limitations: Each node merging will lead to the loss and averaging of local structure and node features, especially after multiple iterations of coarsening this local operation is more likely to lead to the loss of fine-grained structure information of the graph and the weakening of the original semantic similarity between nodes [61]. In data sets such as Cora, the negative effects of multi-scale coarsening are more obvious due to the strong semantic heterogeneity and complex reference relationships of node features, which reduces the model's ability to maintain fine-grained structural features and effectively depict global structures.

Although we have updated the weight of edges through cosine similarity to try to alleviate the information loss in the coarser process, cosine similarity essentially only measures the similarity of local feature Spaces, and cannot effectively capture the deeper, higher-order or global structural dependencies between nodes. For example, in a Cora dataset, the textual features of two papers may show local similarity, but their citation networks may be in different subject areas or knowledge communities, and this global structural difference is difficult to accurately reflect through local feature similarity (such as cosine similarity). Therefore, in the process of coarsening nodes gradually, the potential implicit semantic association between these nodes is easy to be ignored or incorrectly merged, which leads to the connectivity of node clustering and the accuracy of module partitioning. In addition, the use of fixed multi-scale fusion weights alone may not be able to effectively adapt to the feature importance differences of different scales in Cora data sets, for example, some scales may be more critical to identify specific topics, resulting in inappropriate information loss in the process of information fusion, and further weakening the discriminant ability of the clustering model [62].

On the other hand, our model relies on the dynamic selection strategy of positive and negative samples in contrast learning, and the sample selection process is mainly based on the cluster assignment of the current node features. However, in Cora data sets, node features often contain rich semantics and show certain intra-class diversity. Simply dividing positive and negative samples based on node features and static clustering may be difficult to effectively avoid the "False-negative sample" problem. That is, nodes with similar semantics but different category labels may be wrongly classified as negative samples, which significantly interferes with the quality of contrast learning and the model's ability to discriminate complex structures. Therefore, in our future research work, we should focus on introducing dynamic and semantically sensitive contrast learning positive and negative sample selection methods to reduce the influence of false negative sample pairs and improve the generalization performance of the model on complex data sets [63].

## 6. Conclusion

This paper introduces the MPCCL model for self-supervised attributed graph clustering, addressing the limitations of existing methods in capturing long-distance dependencies, feature collapse, and information loss. MPCCL integrates multi-scale graph coarsening and contrastive learning techniques to capture the multi-level structural information of graphs, enhancing the discriminative power and robustness of node embeddings and offering an innovative solution for graph clustering tasks. Experimental results on datasets such as ACM, DBLP, Cora, Citeseer and Reuters demonstrate that MPCCL significantly improves on ACC, NMI, ARI, and F1 metrics compared to suboptimal methods, particularly on the ACM dataset, where it achieves improvements of 15.24% and 11.69% on NMI and ARI metrics over the CLAGC method. These results validate the remarkable advantages of MPCCL in improving clustering performance and model robustness, proving its wide adaptability and effectiveness across different types of graph datasets. The functionality of MPCCL is primarily achieved through multi-scale graph coarsening and a one-to-many contrastive learning mechanism. The multi-scale graph coarsening module hierarchically reduces the number of nodes and edges to preserve key graph structural information, while the contrastive learning module enhances the discriminative power of node embeddings by constructing positive and negative sample pairs. Additionally, the self-supervised learning module maintains the continuity and consistency of node embeddings across different scales through graph reconstruction loss and KL divergence, effectively ensuring the integrity of graph structural information. Despite MPCCL's excellent performance in enhancing graph clustering, there is room for improvement in computational complexity, particularly in the multi-scale coarsening and contrastive learning modules, which incur high computational costs on large-scale datasets. Moreover, the current positive and negative sample selection strategy needs further optimization to enhance adaptability to dynamic graph data. Future research will focus on reducing the model's computational complexity to improve MPCCL's applicability on large-scale graph data. Additionally, exploring more flexible positive and negative sample selection strategies could strengthen the model's generalization ability in dynamic graph environments. Another direction is to extend MPCCL to other complex network data fields, such as social network analysis and bioinformatics, to further validate its potential for multi-domain applications.

## Contribution statement

**Binxiong Li:** Conceptualization, Methodology, Software, Validation, Formal analysis, Writing-Original draft, Review & Editing. **Yuefei Wang:** Conceptualization, Methodology, Project administration, Supervision, Review & Editing, Funding acquisition. **Binyu Zhao:** Methodology, Software, Validation, Investigation, Data curation, Writing-Original draft, Review & Editing. **Heyang Gao:** Methodology, Software, Validation, Investigation, Data curation, Writing-Original draft, Review & Editing. **Benhan Yang:** Methodology, Data curation, Validation. **Quanzhou Luo:** Methodology, Data curation, Validation. **Xue Li:** Methodology, Data curation, Validation. **Xu Xiang:** Methodology, Data curation, Visualization. **Yujie Liu:** Data curation, Visualization, Writing-Original draft, Review & Editing. **Huijie Tang:** Data curation, Visualization, Writing-Original draft, Review & Editing.

## Acknowledgment

The authors thank the editors and the anonymous referees for their valuable comments and efforts. This research is supported by the Sichuan Comic and Animation Research Center, Key Research Institute of Social Sciences of Sichuan Province (DM2024013). In addition, the computational platform for this study is supported by the National Supercomputing Center in Chengdu—Chengdu University Branch.

### A. Proof of Theorem 1.

First, according to **Assumption 1**, we partition the node set $V$ into $n'$ disjoint subsets $C_1, C_2, \ldots, C_{n'}$, i.e., $V = \bigsqcup_{i=1}^{n'} C_i$.

**Define the projection matrix $P$:**

According to **Assumption 3**, we define the projection matrix $P \in \mathbb{R}^{n \times n'}$ with elements:

$$P_{un_i} = \begin{cases} \frac{1}{\sqrt{|C_i|}}, & \text{if node } u \in C_i, \\ 0, & \text{otherwise,} \end{cases} \tag{43}$$

This matrix $P$ satisfies $P^\top P = I_{n'}$, where $I_{n'}$ is the $n' \times n'$ identity matrix. Compute the elements of $P^\top P$:

$$(P^\top P)_{ij} = \sum_{u=1}^{n} P_{ui} P_{uj} = \sum_{u \in V} P_{ui} P_{uj} \tag{44}$$

Since $P_{ui}$ is non-zero only when $u \in C_i$, and similarly for $P_{uj}$, and because $C_i \cap C_j = \emptyset$ when $i \neq j$, we have: $(P^\top P)_{ij} = \sum_{u \in C_i \cap C_j} P_{ui} P_{uj} = 0$. When $i = j$, $(P^\top P)_{ii} = \sum_{u \in C_i} \left(\frac{1}{\sqrt{|C_i|}}\right)^2 = \frac{|C_i|}{|C_i|} = 1$. Therefore, $P^\top P = I_{n'}$, where $I_{n'}$ is the $n' \times n'$ identity matrix.

For an undirected graph $G = (V, E, W)$, its Laplacian matrix is defined as $L = D - A$,, where $D$ is the degree matrix and $A$ is the adjacency matrix. Since node merging is based on equivalent partitioning, the Laplacian matrix $L'$, of the coarsened graph $G'$ can also be represented as $L' = D' - B$,, where $D'$ is the degree matrix of the coarsened graph, and $B$ is the adjacency matrix of the coarsened graph. Due to the properties of the equivalent partition, the adjacency matrix $B$ of the coarsened graph can be expressed as $B = P^\top AP$, and the degree matrix $D'$ of the coarsened graph also satisfies $D' = P^\top DP$. Therefore, the Laplacian matrix $L'$ of the coarsened graph is: $L' = D' - B = P^\top DP - P^\top AP = P^\top (D - A)P = P^\top LP$. Now, we attempt to use the Rayleigh quotient to prove the eigenvalue relationship. For a real symmetric matrix $M$ and a non-zero vector $x$, the Rayleigh quotient is defined as:

$$R(M, x) = \frac{x^\top Mx}{x^\top x}. \tag{45}$$

Let $y \in \mathbb{R}^{n'}$ be an arbitrary non-zero vector, and consider $x = Py \in \mathbb{R}^n$. Since $P^\top P = I_{n'}$, we have $y = P^\top x$. Compute the Rayleigh quotient:

$$R(L', y) = \frac{y^\top L' y}{y^\top y} = \frac{(P^\top x)^\top L' (P^\top x)}{(P^\top x)^\top (P^\top x)} = \frac{x^\top P L' P^\top x}{x^\top x}. \tag{46}$$

Since $P^\top P$ is a projection matrix, we have $PP^\top x = x$, thus:

$$R(L', y) = \frac{x^\top L' x}{x^\top x} \Leftrightarrow R(L', y) = R(L', y) = \frac{x^\top (PP^\top) L (PP^\top) x}{x^\top x} = \frac{x^\top L x}{x^\top x} \Leftrightarrow R(L', y) = R(L, x) \tag{47}$$

Since $x$ is in $\mathbb{R}^n$ and $y$ is in $\mathbb{R}^{n'}$, and $x = Py$, this indicates that for any vector $y$ in the coarsened graph, its Rayleigh quotient matches that of the corresponding vector $x$ in the original graph.

Let $\lambda_k$ and $\lambda'_k$ be the $k$-th eigenvalues of $L$ and $L'$, respectively, arranged in non-decreasing order. According to the extremal properties of the Rayleigh quotient, we have:

$$\begin{aligned} \lambda_k &= \min_{\substack{x \in \mathbb{R}^n \\ x^\top x = 1 \\ x \perp v_1, \dots, v_{k-1}}} x^\top L x, \\ \lambda'_k &= \min_{\substack{y \in \mathbb{R}^{n'} \\ y^\top y = 1 \\ y \perp u_1, \dots, u_{k-1}}} y^\top L' y, \end{aligned} \tag{48}$$

where $v_i$ and $u_i$ are the eigenvectors corresponding to $L$ and $L'$, respectively.

Since $x = Py$, and $\text{rank}(P) = n'$, the subspace $\text{Im}(P)$ is an $n'$-dimensional subspace of $\mathbb{R}^n$. Therefore, for any $y \in \mathbb{R}^{n'}$, $x = Py \in \text{Im}(P)$.

Thus:

$$\lambda'_k = y_k^\top L' y_k = (Py_k)^\top L(Py_k) = x_k^\top L x_k, \tag{49}$$

where $x_k = Py_k$. Thus, $\lambda'_k$ is the minimum Rayleigh quotient of $L$ over the subspace $\mathcal{S} = \{x \mid x = Py, y \in \mathbb{R}^{n'}\}$.

Since the minimization in $\mathbb{R}^n$ (for $\lambda_k$) is over a larger space than in $\text{Im}(P)$, we have:

$$\lambda_k \le \lambda'_k \text{ for } k = 1,2,\dots,n' \tag{50}$$

This completes the proof of Theorem 1.

**B. Proof of Theorem 2-1**

Our goal is to compare the condition numbers of the original Laplacian matrix $L$ and the coarsened Laplacian matrix $L^{(s)}$, proving that:

$$\kappa\left(L^{(s)}\right) = \frac{\lambda_{max}\left(L^{(s)}\right)}{\lambda_{min}^{+}\left(L^{(s)}\right)} \le \frac{\lambda_{max}(L)}{\lambda_{min}^{+}(L)} = \kappa(L), \tag{51}$$

where $\lambda_{min}^{+}(L)$ denotes the smallest non-zero eigenvalue of $L$, also known as the algebraic connectivity. To establish this, we need to prove the following two inequalities:

**i)** The smallest non-zero eigenvalue increases:

$$\lambda_{min}^{+}\left(L^{(s)}\right) \ge \lambda_{min}^{+}(L). \tag{52}$$

**ii)** The largest eigenvalue decreases or does not increase:

$$\lambda_{max}\left(L^{(s)}\right) \le \lambda_{max}(L). \tag{53}$$

The Laplacian matrix $L$ is a symmetric positive semi-definite matrix with non-negative eigenvalues, ordered as: $0 = \lambda_1(L) \le \lambda_2(L) \le \dots \le \lambda_n(L) = \lambda_{max}(L)$. Here, $\lambda_2(L)$ represents the algebraic connectivity of the graph, reflecting the overall connectivity of the graph.

**(1) Proof of $\lambda_{min}^{+}(L^{(s)}) \ge \lambda_{min}^{+}(L)$**

For any non-zero vector $f \in \mathbb{R}^n$ with $f^\top \mathbf{1} = 0$, the Rayleigh quotient of the Laplacian matrix is given by:

$$R_L(f) = \frac{f^\top L f}{f^\top f}. \tag{54}$$

The algebraic connectivity can be expressed as:

$$\lambda_2(L) = \min_{f \neq 0, f^\top \mathbf{1}=0} R_L(f). \tag{55}$$

Let $P \in \mathbb{R}^{n \times n^{(s)}}$ be the projection matrix constructed earlier. For $f^{(s)} \in \mathbb{R}^{n^{(s)}}$ with $f^{(s)\top}\mathbf{1} = 0$, let $f = Pf^{(s)}$. The Rayleigh quotient relation is:

$$R_L(f) = \frac{f^\top L f}{f^\top f} = \frac{f^{(s)\top} P^\top L P f^{(s)}}{f^{(s)\top} P^\top P f^{(s)}} = \frac{f^{(s)\top} L^{(s)} f^{(s)}}{f^{(s)\top} f^{(s)}} = R_{L^{(s)}}\left(f^{(s)}\right). \tag{56}$$

From the coarsened graph to the original graph:

$$\lambda_2\left(L^{(s)}\right) = \min_{f^{(s)} \neq 0, f^{(s)\top}\mathbf{1}=0} R_{L^{(s)}}\left(f^{(s)}\right) = \min_{f \in \mathrm{Im}(P), f \neq 0, f^\top \mathbf{1}=0} R_L(f). \tag{57}$$

Since $\mathrm{Im}(P) \subseteq \mathbb{R}^n$, we have:

$$\lambda_2\left(L^{(s)}\right) = \min_{f \in \mathrm{Im}(P), f \neq 0, f^\top \mathbf{1}=0} R_L(f) \geq \min_{f \neq 0, f^\top \mathbf{1}=0} R_L(f) = \lambda_2(L). \tag{58}$$

Therefore,

$$\lambda_{min}^{+}\left(L^{(s)}\right) = \lambda_2\left(L^{(s)}\right) \geq \lambda_2(L) = \lambda_{min}^{+}(L). \tag{59}$$

**(2) Proof of $\lambda_{max}(L^{(s)}) \leq \lambda_{max}(L)$**

The maximum eigenvalue of the Laplacian matrix can be expressed as:

$$\lambda_{max}(L) = \max_{f \neq 0} R_L(f) = \max_{f \neq 0} \frac{f^\top L f}{f^\top f}. \tag{60}$$

From the original graph to the coarsened graph:

$$\lambda_{max}(L) = \max_{f \neq 0} R_L(f) \geq \max_{f = Pf^{(s)}, f^{(s)} \neq 0} R_L(f) = \max_{f^{(s)} \neq 0} R_{L^{(s)}}\left(f^{(s)}\right) = \lambda_{max}\left(L^{(s)}\right) \Leftrightarrow \lambda_{max}\left(L^{(s)}\right) \leq \lambda_{max}(L) \tag{61}$$

Since we have shown that $\lambda_{min}^{+}(L^{(s)}) \geq \lambda_{min}^{+}(L)$ and $, \lambda_{max}(L^{(s)}) \leq \lambda_{max}(L)$, it follows that:

$$\kappa\left(L^{(s)}\right) = \frac{\lambda_{max}\left(L^{(s)}\right)}{\lambda_{min}^{+}(L^{(s)})} \leq \frac{\lambda_{max}(L)}{\lambda_{min}^{+}(L)} = \kappa(L). \tag{62}$$

Thus, Theorem 2-1 is proven.

**Remark 4.** The reduction in the condition number indicates a more concentrated spectral distribution of the Laplacian matrix, facilitating more effective learning of node representations by graph neural networks and enhancing the model's discriminative power. If Theorems 2-1 and 2-2 hold, then: (1) the smallest non-zero eigenvalue of the coarsened Laplacian matrix is no less than that of the original Laplacian matrix, i.e., $\lambda_{min}^{+}(L^{(s)}) \geq \lambda_{min}^{+}(L)$. This indicates an increase in the algebraic connectivity of the coarsened graph, strengthening its overall connectivity. (2) The largest eigenvalue of the coarsened Laplacian matrix is no greater than that of the original, i.e., $\lambda_{max}(L^{(s)}) \leq \lambda_{max}(L)$. This implies a narrower spectral range for the coarsened graph. Consequently, the condition number of the coarsened Laplacian matrix does not exceed that of the original, i.e., $\kappa(L^{(s)}) \leq \kappa(L)$. This has a positive impact on both the training and performance of graph neural networks. Through a complete proof of the condition number, we further validate the effectiveness of the graph coarsening method based on high spectral similarity when using only the coarsened graph structure. This coarsening approach not only preserves the spectral properties of the original graph but also improves the condition number of the Laplacian matrix, thereby enhancing the numerical stability and learning capability of graph neural networks.

**C. Proof of Theorem 2-2.**

Define the mapping from the original nodes to the coarsened nodes. For a merged node $n_i \in V^{(s)}$, it is formed by merging the original node set $C_i \subset V$.

Projection matrix $P$: defined as $P_{uj} = \begin{cases} \frac{1}{\sqrt{|C_j|}} & \text{if} u \in C_j, \\ 0 & \text{otherwise.} \end{cases}$ where $C_j$ is the node set corresponding to the coarsened node $n_j$. The constructed $P$ satisfies: (1) Orthogonality. $P^\top P = I_{n^{(s)}}$, where $I_{n^{(s)}}$ is the $n^{(s)} \times n^{(s)}$ identity matrix. (2) Projection property. $PP^\top$ is a symmetric projection matrix that projects $\mathbb{R}^n$ onto the coarsened subspace. We define $\tilde{L} = PL^{(s)}P^\top$. Our goal is to estimate $\| L - \tilde{L} \|_2$, which is the spectral norm difference between matrices $L$ and $\tilde{L}$. To compute $L - \tilde{L}$, we need to understand the structure of the Laplacian matrix. The original Laplacian matrix $L = D - W$, and similarly, $L^{(s)} = D^{(s)} - W^{(s)}$. First, consider an arbitrary vector $x \in \mathbb{R}^n$, then

$$x^\top Lx = \sum_{(u,v)\in E} w_{uv}(x_u - x_v)^2. \tag{63}$$

$$x^\top \tilde{L}x = x^\top PL^{(s)}P^\top x = y^\top L^{(s)}y = \sum_{(i,j)\in E^{(s)}} w_{ij}^{(s)}(y_i - y_j)^2, \tag{64}$$

where $y = P^\top x \in \mathbb{R}^{n^{(s)}}$, $y_i = \frac{1}{\sqrt{|C_i|}}\sum_{u\in C_i} x_u$.

$$x^\top(L - \tilde{L})x = \sum_{(u,v)\in E} w_{uv}(x_u - x_v)^2 - \sum_{(i,j)\in E^{(s)}} w_{ij}^{(s)}(y_i - y_j)^2. \tag{65}$$

To further analyze, we partition the edge set $E$ of the original graph into two parts: **i)** Intra-cluster edge set (edges within the same cluster $C_i$): $E_{\text{intra}} = \{(u,v) \in E \mid u, v \in C_i \text{ for some } i\}$. **ii)** Inter-cluster edge set (edges between different clusters): $E_{\text{inter}} = \{(u,v) \in E \mid u \in C_i, v \in C_j, i \neq j\}$.

Therefore,

$$x^\top Lx = \sum_{(u,v)\in E_{\text{intra}}} w_{uv}(x_u - x_v)^2 + \sum_{(u,v)\in E_{\text{inter}}} w_{uv}(x_u - x_v)^2. \tag{66}$$

Similarly, since the coarsening process merges each cluster $C_i$ into a single node, intra-cluster edges are "folded" in

$\tilde{L}$, and only inter-cluster edges are retained. Thus,

$$x^\top \tilde{L} x = \sum_{(i,j)\in E^{(s)}} w_{ij}^{(s)} (y_i - y_j)^2 = \sum_{(u,v)\in E_{\text{inter}}} w_{uv}(x_u - x_v)^2, \tag{67}$$

since $w_{ij}^{(s)} = \sum_{u\in C_i} \sum_{v\in C_j} w_{uv}$, and $y_i = \frac{1}{\sqrt{|C_i|}} \sum_{u\in C_i} x_u$.

$$x^\top (L - \tilde{L}) x = \sum_{(u,v)\in E_{\text{intra}}} w_{uv}(x_u - x_v)^2. \tag{68}$$

Then, we estimate the upper bound of the difference. Since the difference is contributed only by intra-cluster edges, we need to estimate this term. Considering the similarity of node features: (1) High similarity. For $(u, v) \in E_{\text{intra}}$, nodes $u$ and $v$ have highly similar features, so $x_u \approx x_v$. (2) Small feature difference. Assume there exists $\eta > 0$ such that for all $(u, v) \in E_{\text{intra}}$, $|x_u - x_v| \le \eta \parallel x \parallel_2$. (3) Large edge weights. Due to the high similarity of node features, the edge weight $w_{uv}$ is also large.

Estimating the difference:

$$x^\top (L - \tilde{L}) x = \sum_{(u,v)\in E_{\text{intra}}} w_{uv}(x_u - x_v)^2 \le \sum_{(u,v)\in E_{\text{intra}}} w_{uv}(\eta \parallel x \parallel_2)^2 = (\eta \parallel x \parallel_2)^2 W_{\text{intra}}, \tag{69}$$

where $W_{\text{intra}} = \sum_{(u,v)\in E_{\text{intra}}} w_{uv}$.
According to the definition of the matrix spectral norm:

$$\parallel L - \tilde{L} \parallel_2 = \max_{x\neq 0} \frac{x^\top (L - \tilde{L}) x}{\parallel x \parallel_2^2} \le \eta^2 W_{\text{intra}}. \tag{70}$$

Therefore, as long as $\eta^2 W_{\text{intra}} \le \epsilon$, we have:

$$\parallel L - \tilde{L} \parallel_2 \le \epsilon. \tag{71}$$

Thus, Theorem 2-2 is proved.

**Discussion on $\eta$ and $W_{\text{intra}}$.**

Due to the high similarity in features of nodes within the cluster, $x_u \approx x_v$, so $\eta$ is very small. $W_{\text{intra}}$ is the total weight of intra-cluster edges. Although the edge weights are relatively large, the product $\eta^2 W_{\text{intra}}$ can still be controlled within the range of $\epsilon$ because $\eta^2$ is very small. To further analyze the eigenvalue differences between $L$ and $\tilde{L}$, we can use Weyl's inequality.

**Weyl's Inequality**: For symmetric matrices $A$ and $B$, with their eigenvalues ordered in ascending sequence as $\lambda_1(A) \le \lambda_2(A) \le \cdots \le \lambda_n(A)$ and $\lambda_1(B) \le \lambda_2(B) \le \cdots \le \lambda_n(B)$, it holds that:

$$|\lambda_i(A) - \lambda_i(B)| \le \parallel A - B \parallel_2. \tag{72}$$

Since $\parallel L - \tilde{L} \parallel_2 \leq \epsilon$, for all $i$,

$$\left|\lambda_i(L) - \lambda_i(\tilde{L})\right| \leq \epsilon. \tag{73}$$

This implies that the eigenvalues of $L$ and $\tilde{L}$ are very close, meaning their spectral properties are similar. Since node matching is based on high feature similarity, the features of nodes within clusters are very close, making $\eta$ very small. Although $W_{\text{intra}}$ might be relatively large, $\eta^2$ is sufficiently small to ensure $\epsilon$ is within the required bounds. Therefore, the Laplacian matrices $L$ and $\tilde{L}$ are spectrally close, indicating that the coarsened graph $G^{(s)}$ can effectively approximate the spectral properties of the original graph $G$.